\RequirePackage{fix-cm}
\documentclass[twocolumn]{svjour3}          
\smartqed  
\usepackage{graphicx}
\usepackage{amsmath}
\usepackage{amssymb}
\usepackage{subfigure}
\usepackage{algorithm}
\usepackage{algorithmicx}
\usepackage{algpseudocode}
\usepackage{setspace}
\usepackage{longtable}
\usepackage{diagbox}
\usepackage{multirow}
\usepackage{color}

\begin{document}

\title{Category-wise Attack: Transferable Adversarial Examples for Anchor Free Object Detection}


\author{Quanyu Liao \and Xin Wang$^{*}$ \and Bin Kong \and Siwei Lyu \and Youbing Yin \and \\ Qi Song \and Xi Wu$^{*}$
}


\institute{
            Quanyu Liao \and Xi Wu \at
            Chengdu University of Information Technology, China \\
            \email{zankerliao@gmail.com, xi.wu@cuit.edu.cn}
            \and
            Xin Wang \and Bin Kong \and Youbing Yin \and Qi Song \at
            CuraCloud Corporation, Seattle, USA \\
            \email{\{xinw,bink,yin,song\}@curacloudcorp.com}
            \and
            Siwei Lyu
            University at Albany, State University of New York, USA \\
            \email{lsw@cs.albany.edu}
            \and \\
            $^{*}$Corresponding authors: Xin Wang, Xi Wu.
}

\date{Received: date / Accepted: date}

\maketitle
\begin{abstract}
Deep neural networks have been demonstra-\\ted to be vulnerable to adversarial attacks: 
sutle perturbations can completely change the classification results. Their vulnerability 
has led to a surge of research in this direction. However, most works dedicated to attacking 
anchor-based object detection models. In this work, we aim to present an effective and efficient 
algorithm to generate adversarial examples to attack anchor-free object models based on two 
approaches. First, we conduct category-wise instead of instance-wise attacks on the object 
detectors. Second, we leverage the high-level semantic information to generate the adversarial 
examples. Surprisingly, the generated adversarial examples it not only able to effectively attack 
the targeted anchor-free object detector but also to be transferred to attack other object 
detectors, even anchor-based detectors such as Faster R-CNN.
\end{abstract}

\section{Introduction}
%
%
%
%
The development of deep neural network has enabled significant improvement in the performance of many computer vision tasks. However, many recent works show that deep learning-based algorithms 
are vulnerable to adversarial attacks~\cite{carlini2017towards,dong2018boosting,xie2019improving,croce2019minimally,dong2019evading,shi2019curls,xiao2018generating,moosavi2017universal,baluja2017adversarial}: adding imperceptible but specially designed  
adversarial noise can make the algorithms fail. The vulnerability of deep  networks is 
observed in many different problems~\cite{bose2018adversarial,chen2018robust,li2018robust,kurakin2016adversarial,yang2018realistic,tabacof2016adversarial,metzen2017universal,eykholt2017robust,dong2019efficient}, including object detection, one of the most fundamental tasks in computer vision. Adversarial examples 
can help to improve the robustness of neural networks~\cite{liao2018defense,pang2018towards,pang2019improving,athalye2018obfuscated,papernot2016distillation} or to help explain some 
characteristics of the neural network~\cite{dong2017towards}.

Regarding the investigation of the vulnerability of deep models in object detection, previous 
efforts mainly focused on classical anchor-based networks such as Faster-RCNN~\cite{ren2015faster}. 
However, the performance of anchor-based approaches is limited by the choice of the anchor boxes, 
and fewer anchor leads to faster speed but lowers accuracy. Thus, advanced anchor-free models such 
as DenseBox~\cite{huang2015densebox}, CornerNet~\cite{law2019cornernet} and CenterNet~\cite{zhou2019objects} 
are becoming increasingly popular, achieving competitive accuracy with traditional anchor-based 
models with faster speed and stronger adaptability.  However, as far as we know, no existing 
research has been devoted to the investigation of their vulnerabilities. Adversarial attack 
methods for Anchor-based \cite{xie2017adversarial,zhou2019objects} are usually not applicable 
to anchor-free detectors because they generate adversarial examples by selecting the top proposals 
and attacking them one by one.

\begin{figure*}[t]
  \begin{center}
  \includegraphics[width=.49\linewidth]{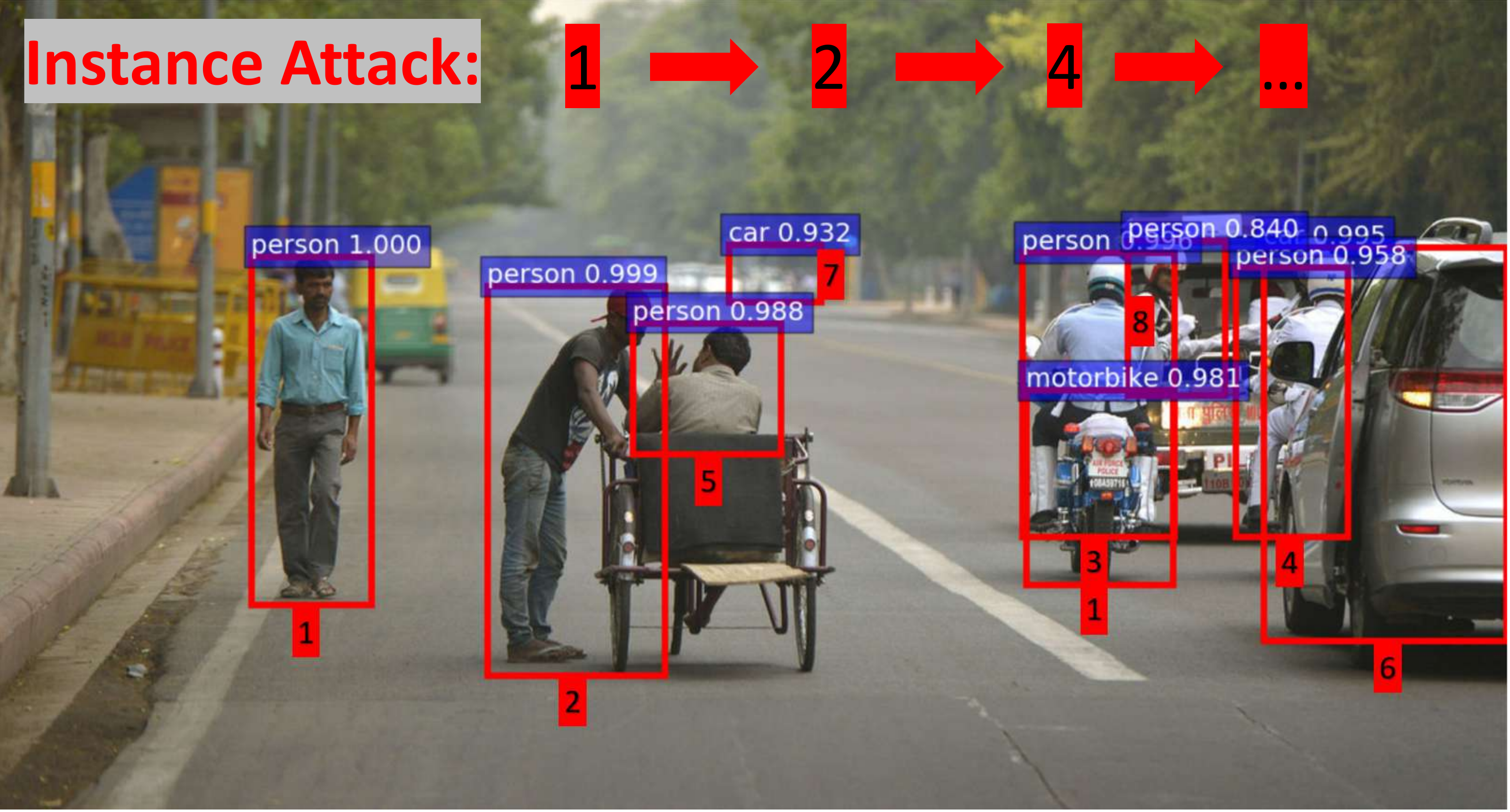}
  \includegraphics[width=.49\linewidth]{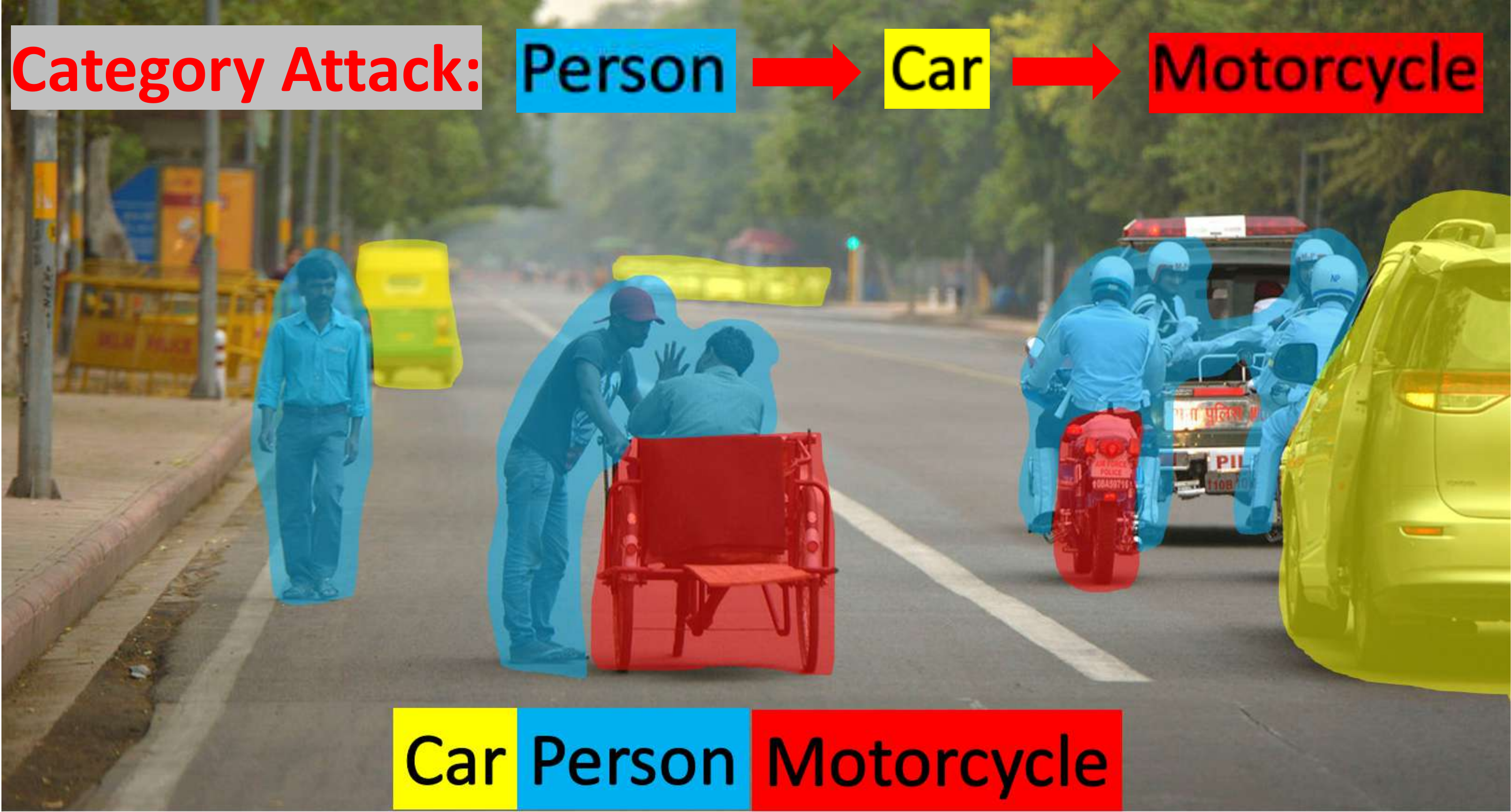}
  \caption{\textbf{Left:} Instance-wise attack. The attacker attacks individual objects one by one. \textbf{Right:} Our category-wise attack algorithm. The attacker attacks all the objects in a category at the same time.}
  \label{TS}
  \end{center}
 \vspace*{-0.5cm}
\end{figure*}

To solve these problems, we propose a new algorithm, \textbf{Category-wise Attack (CA)}, to attack state-of-the-art anchor-free object detection models.
It has modules that jointly attack all the instances in a category (see Fig.~\ref{TS} for details). Unlike most of the existing works~\cite{li2018robust,bose2018adversarial,chen2018robust,xie2017adversarial,zhou2019objects} which use anchor proposal for the attack, we select a set of highly informative pixels in the image according to the heatmap that generated by the anchor-free detector to attack. Those highly informative pixels are the keypoints of detected objects which contains higher-level semantic information for the detector. Additionally, as category-wise attack is challenging, we introduce two {\bf Set Attack} methods based on two popular adversarial attack techniques, which were originally proposed for attacking image classification models by minimizing the $\mathop{L_0}$ and $\mathop{L_\infty}$ perturbations \cite{modas2019sparsefool,goodfellow2014generative}, for attacking object detection models with sparse and dense perturbations, named as Sparse Category-wise Attack (SCA) and Dense Category-wise Attack (DCA) respectably.

To the best of our knowledge, our work is not only the first on attacking anchor-free object detectors but also the first on category-wise attack. When DAG attacking is applied for semantic segmentation that only attack detected segmentation area on the input. Different from DAG, our methods not only attack all detected objects but also attack potential objects that have a high probability to transfer to the rightly detected objects during the attack. Our methods generate adversarial perturbation on larger areas than DAG which can avoid the perturbation overfitting on attacking detected object. Generate perturbation on detected objects and potential objects can bring better-transferring attack performance.

We empirically validated our method on two benchmark datasets: PascalVOC \cite{everingham2015pascal} and MS-COCO \cite{lin2014microsoft}. Experimental results show that our method outperforms the state-of-the-art methods, and the generated adversarial examples achieve superior transferability. To summarize, the contributions of this paper are threefold: 
\begin{itemize}

\item We propose a new method for attacking anchor-free object detection models. Our category-wise attack methods generate more transferable and robust adversarial examples than instance-wise methods with less computational overhead.

\item For two widely used $\mathop{L_p}$ norms, \textit{i.e.} $\mathop{L_0}$ and 
$\mathop{L_\infty}$, we derive efficient algorithms to generate sparse and dense perturbations.

\item Our methods are not only attacking detected pixels but also attacking the potential pixels. Thus, our methods generate perturbation on high semantic information and achieve the best black-box attack performance on two public datasets.

\end{itemize}

Comparing to instance-wise attack methods, our category wise methods focus on global and
high-level semantic information. As a result, they are able to generate more transferable adversarial examples.
Our paper is organized as follows.  We first discuss the related work in section~\ref{sec: related_work}. Then, we provide the details of our methods in section~\ref{sec: approach}. The detailed setup and results of our experiment are described in Section~\ref{sec: experiment}. Finally, we conclude the paper in section~\ref{sec: conclusion}.

\section{Related Work}
\label{sec: related_work}
\noindent \textbf{Object Detection.}  With the fast development of the deep convolutional 
neural networks, many methods have been proposed for object detection~\cite{zhang2020multi,mordan2019end,sun2012object}. We briefly 
divide them into anchor-based and anchor-free approaches. Anchor-based object detection 
models include Faster-RCNN~\cite{ren2015faster}, YOLOv2~\cite{redmon2017yolo9000}, 
SSD~\cite{liu2016ssd}, Mask-RCNN~\cite{he2017mask}, RetinaNet~\cite{lin2017focal}. 
Anchor-based object detectors currently achieve relatively 
higher detection performance.

However, they suffer from two major drawbacks: 1) the network architecture is  
complex and difficult to train, and 2) the scales of the anchors have to be adjusted to adapt to 
specific applications, making it difficult to transfer the same network architecture to 
other dataset/tasks. Recently, anchor-free models like CornerNet~\cite{law2019cornernet}, 
ExtremeNet~\cite{zhou2019bottom} and CenterNet~\cite{zhou2019objects} achieve competitive 
performance with the state-of-the-art anchor-based detectors. These models detect objects 
by extracting object key-points. In contrast to anchor-based detectors, anchor-free methods 
are easier to train and free from issues of scale variance.

\noindent \textbf{Adversarial Attack for Image Classification.} 
Goodfellow \emph{et al.}~\cite{goodfellow2014explaining,szegedy2013intriguing} first showed 
that deep neural networks are vulnerable to adversarial examples: adding deliberately generated imperceptible perturbations to the input images can make the deep neural networks output totally wrong results. Since then, a lot of efforts have been devoted to this line of research~\cite{croce2019scaling,goswami2019detecting}. To be perceptually imperceptible, most existing adversarial attack algorithms aim to minimize $\mathop{L_p}$ norm of the adversarial perturbations. $\mathop{L_2}$ and $\mathop{L_{\infty}}$ norms are two of the most widely used. 

Among all the attack algorithms, two most effective schemes are FGSM~\cite{goodfellow2014generative} and PGD~\cite{madry2017towards}. The procedure of FGSM as follow. First, the gradient of the loss with regard to the input image is computed. Then, the perturbation is assigned with the maximum allowable value in the direction of the gradient. Instead of taking a single large step in FGSM, PGD iteratively takes smaller steps in the direction of the gradient. Compared with FGSM, PGD achieves higher attack performance and generates smaller perturbations. To generate even smaller perturbations, Deepfool was proposed in~\cite{moosavi2016deepfool}. It uses the hyperplane to approximate the decision boundary and iteratively computes the lowest euclidean distance between the input image and the hyperplane to generate the perturbations. 

However, these methods suffer from one problem: the generated perturbations are extremely dense--these algorithms have to change almost every pixel of the input image. As a result, the $\mathop{L_0}$ norm of the perturbation is extremely high. To address this issue, some sparse attack algorithms such as JSMA~\cite{papernot2016limitations}, Sparse Fool, and One-Pixel Attack~\cite{su2019one} are proposed to generate sparser perturbations.

\noindent \textbf {Adversarial Attack for Objection Detection.} 

The research on the vulnerability of object detection models is scarce. 
Xie \emph{et al.}~\cite{xie2017adversarial} revealed that anchor-based object detection algorithms are susceptible to  white-box attacks. UEA~\cite{zhou2019objects} leveraged generative adversarial networks (GANs) to improve the transferability of adversarial examples. While DAG and UEA achi-\\eved the SOTA attack performance on anchor-based detectors, they suffer from three shortcomings: \textbf{(1)} DAG and UEA are based on instance-wise attacks, which only attack one object at a time. As a result, the process is extremely inefficient especially when there exist many objects in the input image. \textbf{(2)} UEA relies on a separate GAN network, which needs to be trained each time before being used on a different dataset. Therefore, it is more complex to apply it to a new problem/dataset than optimization-based methods. \textbf{(3)} Adversarial examples generated by DAG show poor transferability. While anchor-free object detection models are becoming increasingly popular, no existing research has been devoted to attacking these anchor-free detectors. Notably, DAG and UEA can not be applied to attack anchor-free models as they attack detectors by attacking the top proposals and anchor-free detectors do not rely on proposals. Similarly, other anchor proposal-based attackers~\cite{li2018robust,bose2018adversarial,chen2018robust} 
are also not applicable.

\section{Approach}
\label{sec: approach}
Anchor-free object detectors take as input an image and yield a heatmap for each object category. Then, the objects in the bounding boxes of each category are extracted from the corresponding heatmap with non-maximum suppression. This motivates us to propose category-wise attack instead of instance-wise attack. In this section, we introduce the details of the proposed Category-wise Attack (CA) algorithms. We first formally formulate our category-wise attack problem in Section \ref{Problem_Formulation}. Then, in Section \ref{Pixels_Selection}, we introduce the category-wise target pixel set selection procedure, which is used in both SCA and DCA. Finally, we propose SCA and DCA to generate sparse and dense perturbations respectively in Section \ref{sec: sca} and \ref{sec: dca}. 

\begin{figure}[t]
\begin{center}
\includegraphics[width=3.3in]{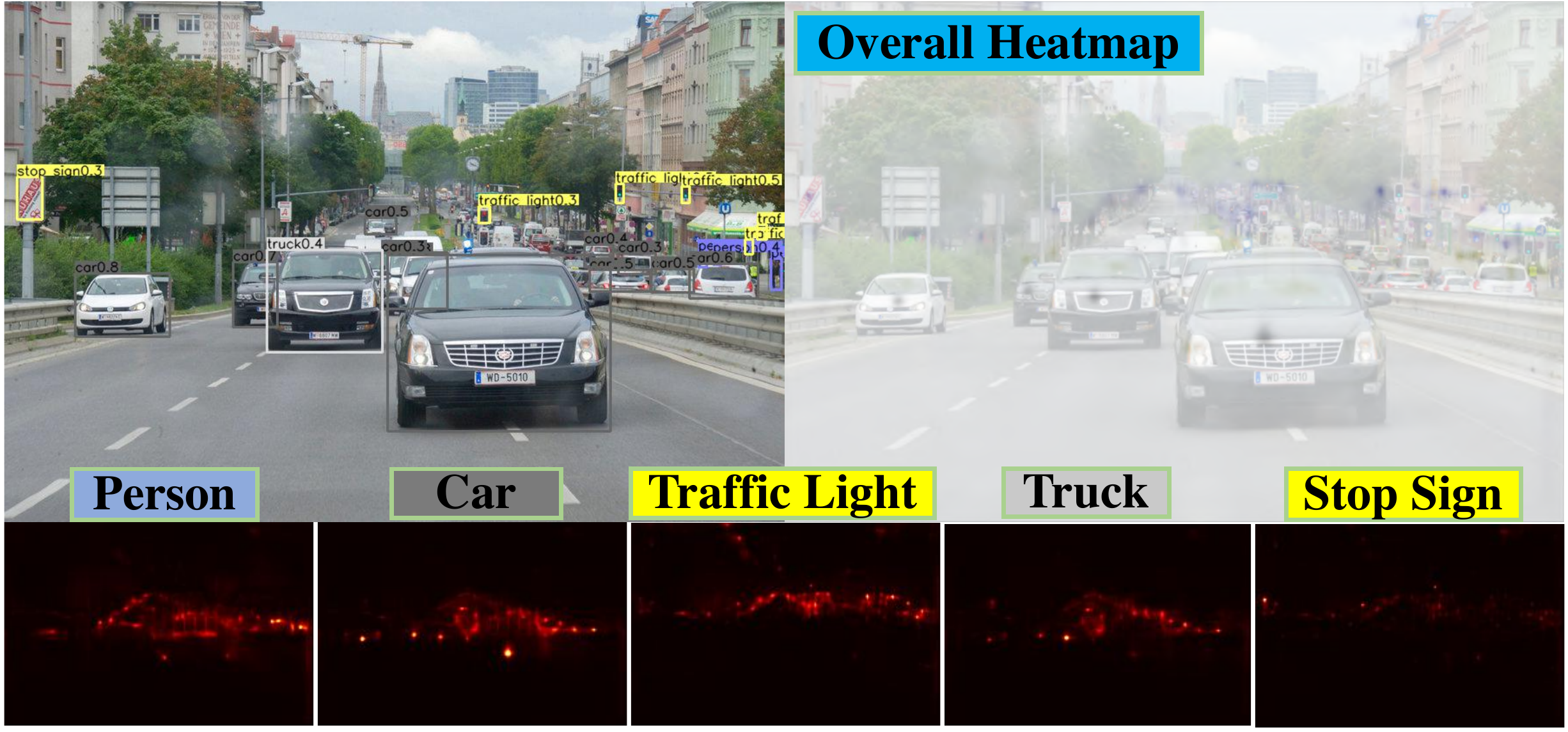}
\end{center}
\vspace*{-0.3cm}
\caption{\textbf{First row:} Detected results and the overall heatmap of CenterNet \cite{zhou2019objects}. \textbf{Second row:} Selected target pixels (Red) for each category.}
\label{fig_pix_select}
\vspace*{-0.5cm}
\end{figure}

\subsection{Problem Formulation}
\label{Problem_Formulation}

In this section, we formally define the category-wise attack problem for anchor-free detectors. Suppose there exist $K$ object categories which include the detected object instances, $C_k$ ($k=1,2,...,K$). We denote the pixels of all the detected object instances in category $C_k$ on the original input image $x$ as target pixel set $P_k$. And $p$ denotes a detected pixel of an object instance.  Specifically, it is formulated as a constrained optimization problem:

\begin{equation}
   \begin{aligned}
       \mathop{\text{minimize}} \limits_{r} \quad & \Vert{r} \Vert_{p} \\
       s.t. ~ \quad & ~\forall k, p\in P_k \\
       & \text{argmax}_n\{f_n(x+r, p)\} \not = C_k \\
       & t_{min} \leq x+r \leq t_{max}
   \end{aligned}
   \label{categorywiseoptimization}
\end{equation}

where $r$ is the adversarial perturbation. $\Vert{\cdot}\Vert_{p}$ is the $L_p$ norm, with $p$ possibly be $0$, $1$, $2$, and ${\infty}$. $x$ denotes the clean input. $x+r$ denotes the adversarial example. $f(x+r, p)$ denotes the classification score vector~(logistic) and the $f_n(x+r, p)$ denotes the $n^{th}$ value of the score vector.
${\arg\max}_n\{f_n(x+r, p)\})$ denotes the predicted object category on a detected pixel $p$ of object of adversarial example $x+r$. $\mathop{t_{max}}$, $\mathop{t_{min}}$  $\mathop{\in}\mathbb{R}^n$ are the maximum and minimum allowable pixel value, which constrains $r$.

In this work, we approximate $P_k$ with the heatmap of category $C_k$ generated by CenterNet \cite{zhou2019objects}, an anchor-free detector. The details are described in section \ref{Pixels_Selection}.

\subsection{Category-wise Target Pixel Set Selection}
\label{Pixels_Selection}

The first step of our methods is to generate the target pixel set to attack, which is composed of the detected pixels and the potential pixels (details in the third paragraph) from the heatmap of the CenterNet.

The attacking procedure of anchor-based attackers is equivalent to using all the available proposals to attack the targeted object instance. Such a mechanism changes non-informative proposals, such as some proposals covering the whole image or very small scope of the image, leading to a higher computational overhead. Instead of attacking all available proposals, we leverage the heatmap, which contains higher-level semantic information for the detectors.
To illustrate this procedure, we use Fig.~\ref{fig_pix_select} as an example. Fig.~\ref{fig_pix_select} shows the detected results, overall heatmap of the CenterNet \cite{zhou2019objects}, and the selected target pixels for each category. As is shown in Fig.~\ref{fig_pix_select}, we divide the detected pixels on the overall heatmap into target pixel set $\{P_1, P_2, ..., P_k\}$.

\begin{figure}[t]
    \begin{center}
    \includegraphics[width=3.3in]{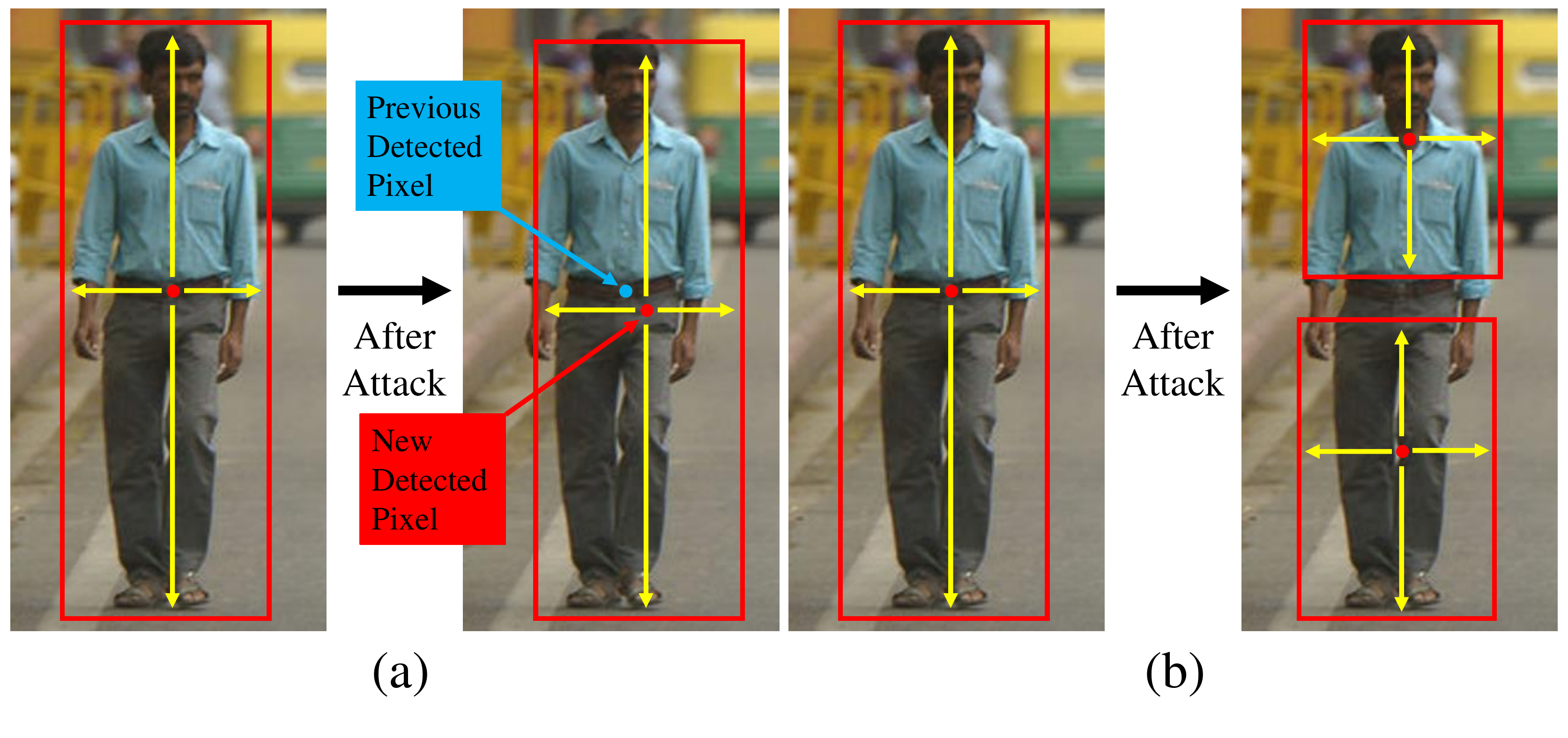}
    \end{center}
    \vspace*{-0.3cm}
    \caption{
    \textbf{Red} points denote the detected pixels.
    \textbf{Blue} point denotes the attacked pixels.
    \textbf{(a)-Left \& (b)-Left:} The detected pixel is centered in the center of the person.
    \textbf{(a)-Right:} Result of only attacking the detected pixels. The neighbor pixel of the previously detected pixel is detected as the correct object by the detector during the attack. This makes IOU not significantly lower.
    \textbf{(b)-Right:} Result of only attacking the detected pixels. The center of the top half and the bottom half of the person appear new detected pixels which still detected as person. Which is also makes IOU not significantly lower.}
    \label{select_pixel}
    \vspace*{-0.5cm}
\end{figure}

In principle, after attacking all detected pixels, the CenterNet should not detect any object on the adversarial example which closes to the origin input. But the CenterNet still detect the same object after attacking all detected pixels, similar situation also happened in DAG \cite{xie2017adversarial}. We check up the heatmap and make sure all detected pixels are changed to the incorrect category and find two reasons that why he CenterNet still work.

\begin{itemize}
\item The neighbor pixels are changed to the previous correct category and be detected as the correct object by the detector. Such as some neighbor background pixels' confidence levels are increased that make themselves become the detected pixel with the correct category. Because the newly detected pixels' location is near to the previously detected pixels, so the newly detected object has the same category as the old object, and the detected box of the newly detected object is close to the old detected object. See Fig.~\ref{select_pixel}-(a).

\item The CenterNet regards the center pixels of the object as the key-points. So in some cases, after attacking the detected pixels which are centered in the center of the object, newly detected pixels appear in the other position of the object. An example is shown in Fig.~\ref{select_pixel}-(b).
\end{itemize}

We call the pixels that may produce the above two changes as \textbf{potential pixels}. These two effects make the CenterNet capable of detecting the correct object on the adversarial example. To solve these two effects, our method not only attacking the detected pixels but also attacking the potential pixels.

In order to attack detected pixels and potential pixels at the same time, the target pixel set $P_k$ is decided by setting an attack threshold $t_{attack}$. The pixels whose probability score in the heatmap is above the $t_{attack}$ are regarded as target pixels and attacked. Setting $t_{attack}$ can help us construct $P_k$ that include detected pixels and potential pixels.
The process for constructing $P_k$ is summarized as follows:

\begin{equation}
   \begin{aligned}
        & P_k = \{ p~\vert~p > t_{attack}\} \\
        s.t. ~ \quad & p \in \text{\emph{heatmap}} \\
        & \text{argmax}_n\{f_n(x+r, p)\} = C_k
    \end{aligned}
\end{equation}
where the $heatmap$ denotes the heatmap of the CenterNet and $heatmap \in [0, 1]^{\frac{W}{R}\times\frac{H}{R}\times n}$. $W$ and $H$ denote the width and height of the input image. $R$ denotes the output stride of the CenterNet and the $n$ is the number of pixel category of the heatmap.

The weakly supervised attack can also help the attack algorithm to leverage the heatmap to perceive global and higher-level category-wise semantic information, and also reduce the computation time by avoiding attack the instance one by one. We use the target pixel set $P$ as the input for both SCA and DCA.

\begin{figure}[t]
   \begin{center}
   \includegraphics[width=.9\linewidth]{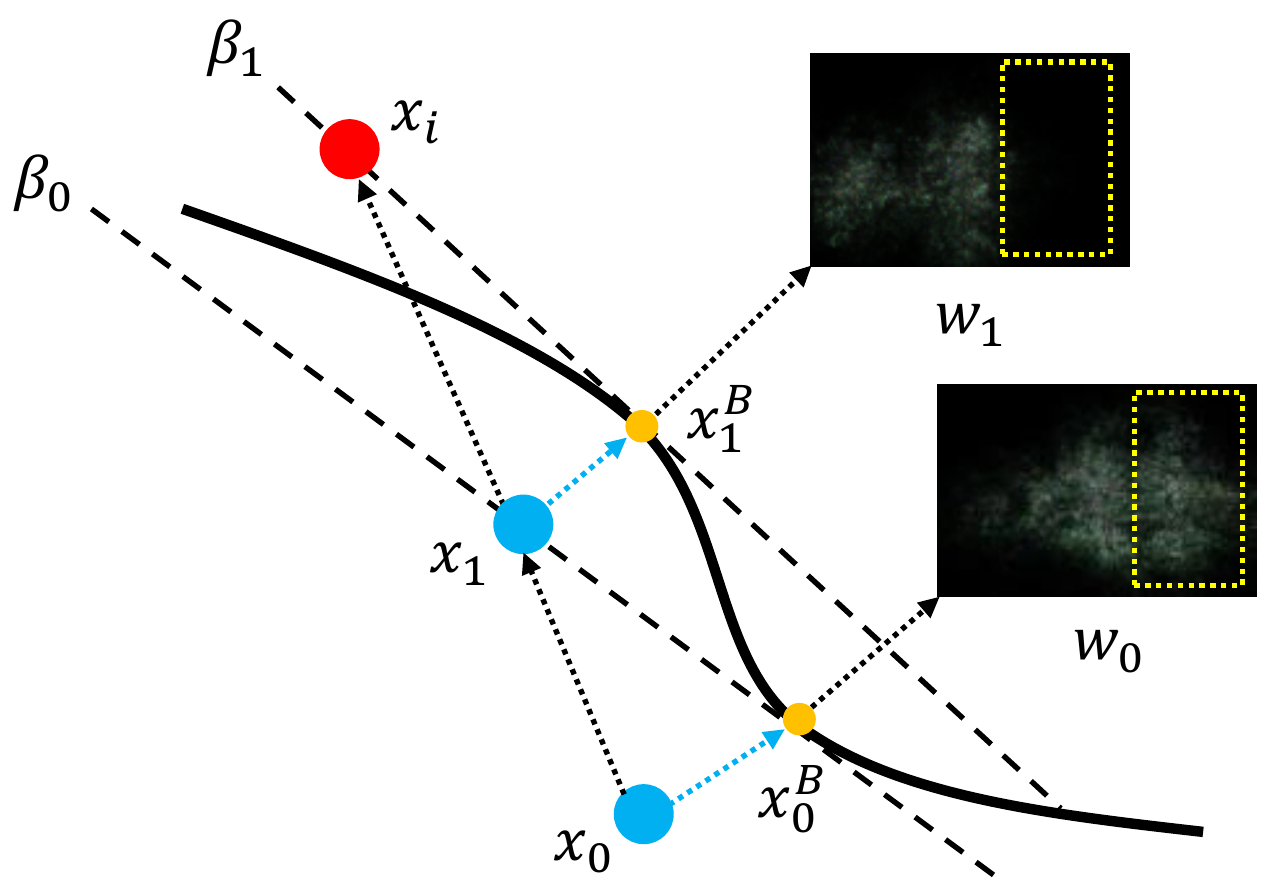}
   \end{center}
   \vspace*{-0.3cm}
   \caption{Illustration of SCA, we use the `Car' category in Fig. \ref{fig_pix_select} as an example. The dash lines are the approximated linear decision boundaries in two consecutive iterations. The two images on the right side are the visualization of the normal vector $\mathop{w}$. We can see that the weights for the `Car' object are reduced (yellow box).}
   \label{fig_sca_car}
   \vspace*{-.5cm}
\end{figure}

\subsection{Sparse Category-wise Attack}
\label{sec: sca}

The goal of the sparse attack is to make the detectors generate false prediction while perturbing the minimum number of pixels in the input image. This is equivalent to setting $p=0$ in our optimization problem (\ref{categorywiseoptimization}).

Sparse Fool \cite{modas2019sparsefool} was originally proposed to generate sparse perturbations for the image classification task. It utilizes a linear solver process to achieve lower $\mathop{L_0}$ perturbations. In this section, we illustrate how to integrate the Sparse Fool method into our anchor-free object detection attacking algorithm, which we term as Sparse Category-wise Attack (SCA) in this paper.

The proposed SCA algorithm is summarized in Algorithm \ref{algorithm_sca}. Given an input image $x$ and category-wise target pixels set $\{P_1, P_2, ..., P_k\}$, our algorithm first to select $P_K$ which has highest total probability score to generate the locally target pixel set $P_{\text{target}}$ of the target category. In this process, we compute the total probability score of each category and choose the highest one as the target category, then use all detected pixels and potential pixels that are still not attacked successfully of the target category to construct $P_{\text{target}}$.

\renewcommand{\algorithmicrequire}{\textbf{Input:}}
\renewcommand{\algorithmicensure}{\textbf{Output:}}
\begin{algorithm}[t]
    \setstretch{1.3}
    \caption{Sparse Category-wise Attack (SCA)}
    \label{sca}
    \begin{algorithmic}

        \Require
        image $x$, target pixel set $\{P_1, P_2, ..., P_k\}$, \\
        available categories $\{C_1, C_2, ..., C_k\}$
        \Ensure
        perturbation $r$

        \State{Initialize: $x_1 \leftarrow x, i \leftarrow 1, j \leftarrow 1, P_{(i)} \leftarrow P$}

        \While{$\{P_1, P_2, ..., P_k\} \not\in \varnothing$}
        
            \State$K = \text{argmax}_k\sum_{p\in P_k} softmax_{C_k}~f(x_i, p)$
            
            \State$P_{target, j=1} \leftarrow P_K$

            \State$x_{i,j} \leftarrow x_i$

            \While{$j \leq M_s$ or $P_{target, j} \in \varnothing $}

                \State$x_j^B = \text{DeepFool}~(x_{i,j})$
                \State$w_j = \text{ApproxBoundary}~(x_j^B, P_{\text{target}, j})$
                \State$x_{i, j+1} = \text{LinearSolver}~(x_{i,j},w_j, x_j^B)$
                \State$P_K = \text{RemovePixels}~(x_{i,j}, x_{i,j+1}, P_K)$

                \State$j = j + 1$

            \EndWhile

            \State$x_{i+1} \leftarrow x_{i,j}$
            \State$i = i + 1$

        \EndWhile
 
        \State{\textbf{return} $r = x_i - x_1$ }
        
    \end{algorithmic}
    \label{algorithm_sca}
\end{algorithm}

\renewcommand{\algorithmicrequire}{\textbf{Input:}}
\renewcommand{\algorithmicensure}{\textbf{Output:}}
\begin{algorithm}[t]
    \setstretch{1.3}
    \caption{ApproxBoundary}
    \label{ab}
    \begin{algorithmic}

        \Require
        dense adversarial example $x^B$, locally target pixel set $P_{target}$
        \Ensure
        normal vector $w$

        \State$ w^{'} \leftarrow \nabla\sum_{p_i \in P_{target}}{f_{ argmax_n f_n(x^B, p_i)}(x^B, p_i)}$ 
        \State$ \quad\quad -\nabla\sum_{p_i \in 
        P_{target}}{f_{ argmax_n f_n(x, p_i)}(x^B, p_i)}$
        \State$ w \leftarrow \frac{w^{'}}{\vert w^{'}\vert}$
        
        \State{\textbf{return} $w$ }
        
    \end{algorithmic}
\end{algorithm}

\renewcommand{\algorithmicrequire}{\textbf{Input:}}
\renewcommand{\algorithmicensure}{\textbf{Output:}}
\begin{algorithm}[t]
    \setstretch{1.3}
    \caption{Category-Wise DeepFool}
    \label{df}
    \begin{algorithmic}

        \Require
        image $x$, locally target pixel set $P_{target}$
        \Ensure
        dense adversarial example $x^B$

        \State{Initialize: $x_0 \leftarrow x, i \leftarrow 0, P_0 \leftarrow P_{target}$}

        \While{$P_{i} \cap P_{target} \not= \varnothing$}
        
            \For{$k \not= target$}

                \State$w_k \leftarrow \nabla\sum_{p_j \in P_i}{f_{k}(x_i, p_j)}$
                \State ~~\quad\quad$-\nabla\sum_{p_j \in P_i}{f_{target}(x_i, p_j)}$
                
                \State$score_k \leftarrow \sum_{p_j \in P_i}f_k(x_i, p_j)$
            
            \EndFor
            
            \State$l \leftarrow \text{argmin}_{k\not=target}~\frac{\vert score_k\vert}{\Vert w_k\Vert_2}$
            \State$pert_i \leftarrow \frac{\vert score_l\vert}{\Vert w_l\Vert_2^2}w_l$
            \State$x_{i+1} \leftarrow x_i + pert_i$
            \State$P_{i+1} \leftarrow \text{RemovePixels}~(x_{i}, x_{i+1}, P_i)$
            \State$i \leftarrow i + 1$
        \EndWhile
 
        \State{\textbf{return} $x^B \leftarrow x_{i+1}$ }
        
    \end{algorithmic}
\end{algorithm}

Next, we generate sparse perturbation by minimizing  $\Vert{r}\Vert{}_0$ 
according to $P_{\text{target}}$. Unfortunately, this is an NP-hard problem. We adopt the method employed by DeepFool \cite{moosavi2016deepfool} and SparseFool~\cite{modas2019sparsefool} to relax this problem by iteratively approximating the classifier as a local linear function. However, these methods are proposed for image classification. To satisfied the category-wise attack on the object detector, we propose \textit{Category-Wise DeepFool (CW-DF)} (See Algorithm.\ref{df}) to generate an initial adversarial example $x^B$ at first. More specifically,  CW-DF computes perturbation on the $P_{target}$, which is different from the original DeepFool computes perturbation on the classifier's output.
Then, the SCA using the \textit{ApproxBoundary} (see Algorithm.~\ref{ab}) to approximate the decision boundary. The decision boundary is locally approximated with a hyperplane $\beta$ passing through $x^B$:

\begin{equation}
   \begin{aligned}
       & \beta \ \mathop{=}^{\triangle} \ \{x:w^T(x - x^B) = 0 \}
    \end{aligned}
\end{equation}
where $w$ is the normal vector of the hyperplane $\beta$, which is approximated by the following equation~\cite{modas2019sparsefool}:
\begin{equation}
   \begin{aligned}
       w :=~&\nabla\sum_{i=1}^n{f_{\text{argmax}_n{f_n(x^B, p)}}(x^B, p)} \\
       &- \nabla\sum_{i=1}^n{f_{\text{argmax}_n{f_n(x, p)}}(x^B, p)}
\end{aligned}
\end{equation}
Then, a sparser adversarial perturbation can be computed by the LinearSolver process~\cite{modas2019sparsefool} and generate the adversarial perturbation. An illustration of the key part of the attack is given in Fig.~\ref{fig_sca_car}.

To satisfy the problem~(\ref{categorywiseoptimization}), the pixel intensity of adversarial example $x_{i+1} = x_{i} + pert$ should clipped in $[0,~255]$, where $pert$ denotes the adversarial perturbation. To better control the imperceptibility of our method, we constrain the pixel intensity of $x_{i} + pert$ to lie in the interval $\pm\epsilon_S$ around the clean input $x_{0}$, where the $\epsilon_S$ denotes the pixel clipped value. The value range of the $\epsilon_S$ is $[0,~1]$.
\begin{equation}
    \begin{aligned}
    0 \leq x_{0} - \epsilon_S\times255 \leq x_i + pert \leq x_{0} + \epsilon_S\times255 \leq 255
    \end{aligned}
\end{equation}
Different $\epsilon_S$ will lead to different attack performance and the imperceptibility of our method. 

Our algorithm alternates between $RemovePixels$~\ref{rp} and generating $P_{target}$ by selecting different $P_k$. After attacking $P_{\text{target}}$ in the each inner loop, we use \textit{RemovePixels} to update $\mathop{P_K}$, which takes as input $x_{i,j}$, $x_{i,j+1}$ and $P_k$. Specifically, the step of \textit{RemovePixels} first generates a new heatmap for the perturbed image $x_{i,j+1}$ with the detector. Then, \textit{RemovePixels} checks whether the probability score of the pixels on $P_K$ is still higher than the $t_{attack}$ on the new heatmap. The pixels with higher probability score than the $t_{attack}$ are retained, and pixels with lower probability score is removed from $P_K$, generating an updated target pixel set $P_K$. If $\{P_1, P_2, ..., P_k\} \in \varnothing$, the attack for all objects of $x$ is successful and we output the generated adversarial example.

\renewcommand{\algorithmicrequire}{\textbf{Input:}}
\renewcommand{\algorithmicensure}{\textbf{Output:}}
\begin{algorithm}[t]
    \setstretch{1.3}
    \caption{RemovePixels}
    \label{rp}
    \begin{algorithmic}

        \Require
        image $x$, adversarial example $x_{adv}$, target pixel set $P$
        \Ensure
        new target pixel set $P_{new}$

        \State{Initialize: $ i \leftarrow 0, P_{new} \leftarrow \varnothing$}

        \For{$p_i \in P$}
            
            \If{$\text{argmax}_nf_n(x_{adv}, p_i) = \text{argmax}_nf_n(x, p_i)$}
                \State$P_{new} \leftarrow P_{new} \cup p_i$
            \EndIf
            \State{$i \leftarrow i + 1$}
        \EndFor
 
        \State{\textbf{return} $P_{new}$ }
        
    \end{algorithmic}
\end{algorithm}

Note that the SCA will not fall into an endless loop. In some iterations, if SCA fails to attack any pixels of the $P_{target}$ in the inner loop, the SCA will attack the same $P_{target}$ in the next iteration. During this process, SCA keeps accumulating perturbations on these pixels. Thus, those perturbations reduce the probability score of each pixel of the $P_{target}$, until the probability score lower than the $t_{attack}$. After that, $P_{target}$ is attacked successfully.

\begin{figure}[t]
  \begin{center}
  \includegraphics[width=\linewidth]{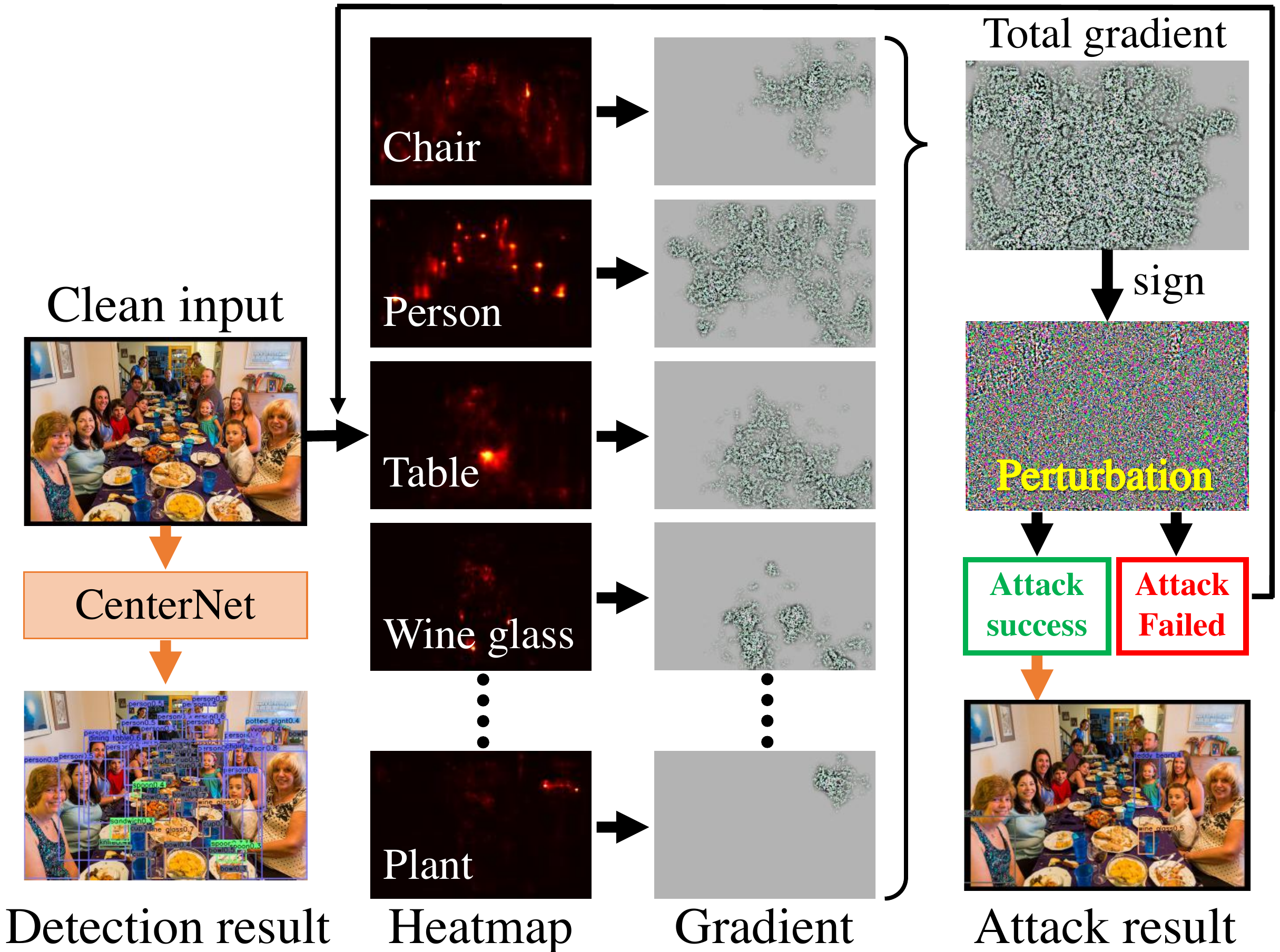}
  \end{center}
  \vspace*{-0.3cm}
  \caption{Illustration of DCA. First, we extract the heatmap for each object category and compute gradient of each object category. After normalizing the adversarial gradients with  $\mathop{L_\infty}$, we add up all gradients and generate adversarial perturbation by further applying $sign$ operation on it. Finally, we check whether the attack is successful. If not, we further generate new perturbation on existing adversarial example in the next iteration.}
  \label{fig_dca_illustration}
\end{figure}

\renewcommand{\algorithmicrequire}{\textbf{Input:}}
\renewcommand{\algorithmicensure}{\textbf{Output:}}
\begin{algorithm}[t]
    \setstretch{1.3}
    \caption{Dense Category-wise Attack (DCA)}
    \label{dca}
    \begin{algorithmic}

        \Require
        image $x$, target pixel set $\{P_1, P_2, ..., P_k\}$\\
        available categories $\{C_1, C_2, ..., C_k\}$, \\
        perturbation amplitude $\epsilon_D$
        \Ensure
        perturbation $pert$

        \State{Initialize: $x_1 \leftarrow x, i \leftarrow 1, j \leftarrow 1, P_{i} \leftarrow P, pert_1 \leftarrow 0$}

        \While{$\{P_1, P_2, ..., P_k\} \not\in \varnothing$ and $i \leq M_D$}

            \State$r_i \leftarrow 0, j \leftarrow 1$

            \While{$j ~\leq= k$}

                \If{$P_{\text{j}} \not= \varnothing$}
                
                    \State$\text{loss}_{\text{sum}} \leftarrow \sum_{p_n\in P_{\text{j}}}\text{CE}~(f(x_i, p_n),~C_j)$
                    \State$r_j \leftarrow \bigtriangledown_{x_i}\text{loss}_{\text{sum}}$
                    \State$r_j' \leftarrow \frac{r_j}{\Vert r_j\Vert_{\infty}}$
                    \State$r_{i} \leftarrow r_{i} + r_j'$

                \EndIf
                
                \State$j \leftarrow j + 1$

            \EndWhile

            \State$pert_i \leftarrow \frac{\epsilon_D}{M_D}\cdot sign(r_{i})$
            \State$x_{i+1} \leftarrow x_i + pert_i$
            \State$\{P_1, ..., P_k\} \leftarrow \text{RemovePixels}~(x_i, x_{i+1}, \{P_1, ..., P_k\})$
            \State$i \leftarrow i + 1$

        \EndWhile

        \State{return $\mathop{pert = x_i - x_0}$}
        
    \end{algorithmic}
    \label{algo: algorithm_dca}
\end{algorithm}

\subsection{Dense Category-wise Attack}
\label{sec: dca}

To further investigate the effect of minimizing $L_{\infty}$ of the perturbation, we set $p=\infty$ in our optimization problem (\ref{categorywiseoptimization}). Previous studies FGSM~\cite{goodfellow2014generative} and PGD~\cite{madry2017towards} are two of the most widely used approaches to attack deep neural networks by minimizing $L_{\infty}$. FGSM generates adversarial examples by taking the maximum allowable step in the direction of the gradient of the loss with regard to the input image. Instead of taking a single large step in FGSM, PGD iteratively takes smaller steps in the direction of the gradient. Compared with FGSM, PGD achieves higher attack performance and generates smaller $L_{\infty}$ perturbations~\cite{madry2017towards}. Therefore, our adversarial perturbation generation procedure is base on the PGD, which generates dense perturbations and named as dense category-wise attack (DCA). 

The DCA is summarized in Algorithm~\ref{algo: algorithm_dca}. 
Given an input image $x$ and category-wise target pixels set $\{P_1,$ $P_2, ..., P_k\}$, DCA only attacks target pixel set $P_j$ for each available category $C_j$ that contains the detected objects. The procedure of DCA is consists of two loops. DCA computes local adversarial gradient $r_j$ for the category $C_j$ which is attacking in the inner loop $j$ and add up all $r_j$ to obtain total adversarial gradient $r_i$ for the all detected categories $\{C_1, C_2, ..., C_k\}$, then generate adversarial perturbation $pert_i$ using $r_i$ in the outer loop $i$.

In each iteration $j$ of the inner loop, DCA computes the gradient for all objects of $P_j$, which is different from instance-wise attack algorithms that only computes the gradient for single object~\cite{xie2017adversarial}. DCA first computes the total loss $loss_{sum}$ of $P_j$ that composed of all detected pixels of each available category $C_j$ as follows:
\begin{equation}
   \begin{aligned}
       \text{loss}_{\text{sum}} = \sum_{p_n\in P_{\text{j}}}\text{CE}~(f(x_i, p_n),~C_j)
\end{aligned}
\end{equation}
where the $CE$ denotes the $CrossEntropy$.

Then, DCA computes local adversarial gradient $r_j$ of the $P_j$ on the $loss_{sum}$ and normalizes it with $L_\infty$, yielding $r_j'$ as follow:
\begin{equation}
   \begin{aligned}
       r_j = & \bigtriangledown_{x_i}\text{loss}_{\text{sum}} \\
       r_j' = & \frac{r_j}{\Vert r_j\Vert_{\infty}}
\end{aligned}
\end{equation}
Finally, DCA adds up all $r_j'$ to generate total adversarial gradient $r_i$.

In each iteration of the outer loop, DCA obtains the total adversarial gradient $r_i$ from the inner loop. Then, in order to accelerate the algorithm, DCA computes perturbation $pert_i$ by applying $sign$ operation to the total adversarial gradient $r_i$ \cite{madry2017towards}:
\begin{equation}
   \begin{aligned}
       pert_i = \frac{\epsilon_D}{M_D}\cdot sign(r_{i})
    \end{aligned}
\end{equation}
where $M_D$ denotes the maximum number of cycles of the outer loop, term $\frac{\epsilon_D}{M_D}$ is optimal max-norm constrained weight to constraint the amplitude of the $pert_i$ \cite{goodfellow2014explaining}. In the final of the outer loop, DCA use $RemovePixels$ (see Section \ref{rp} for details) to refresh the $\{P_1, P_2, ..., P_k\}$ on the $x_{i+1}$.

Note that both SCA and DCA are considered as a non-target multi-class attack method. But the SCA and DCA can easily change to the target attack method.

\section{Experiment}
\label{sec: experiment}
\subsection{Experimental Setup}

\begin{table*}[t]
  
 \begin{center}
     \begin{tabular}{l||c|c||c|c||c|c||c|c}
         \hline
         \multirow{2}{*}{\diagbox{\textbf{From}}{\textbf{To}}} & \multicolumn{2}{c||}{Resdcn18} & \multicolumn{2}{c||}{DLA34} & \multicolumn{2}{c||}{Resdcn101} & \multicolumn{2}{c}{CornerNet} \\ \cline{2-9} 
                                  & mAP           & ATR        & mAP         & ATR      & mAP            & ATR           & mAP            & ATR            \\ \hline \hline
         Clean                    & 0.29          & --         & 0.37        & --       & 0.37           & --            & 0.43           & --             \\ \hline
         R18-DCA                  & 0.01          & 1.00       & 0.29        & 0.21     & 0.28           & 0.25          & 0.38           & 0.12           \\ \hline
         DLA34-DCA                & 0.10          & 0.67       & 0.01        & 1.00     & 0.12           & 0.69          & 0.13           & 0.72           \\ \hline
         R18-SCA                  & 0.11          & 1.00       & 0.27        & 0.41     & 0.24           & 0.57          & 0.35           & 0.30           \\ \hline
         DLA34-SCA                & 0.07          & 0.92       & 0.06        & 1.00     & \textbf{0.09}  & \textbf{0.92} & \textbf{0.12}  & \textbf{0.88}  \\ \hline
     \end{tabular}
 \end{center}

 \caption{Black-box attack results on the MS-COCO dataset. \textbf{From} in the leftmost column denotes the models where adversarial examples generated from.  \textbf{To} in the top row means the attacked models that adversarial examples transfer to.}
 \label{tab3}
\end{table*}

\begin{table*}[t]

 \begin{center}
     \begin{tabular}{l||c|c||c|c||c|c||c|c||c|c}
         \hline
         \multirow{2}{*}{\diagbox{\textbf{From}}{\textbf{To}}} & \multicolumn{2}{c||}{Resdcn18} & \multicolumn{2}{c||}{DLA34} & \multicolumn{2}{c||}{Resdcn101} & \multicolumn{2}{c||}{Faster-RCNN} & \multicolumn{2}{c}{SSD300} \\ \cline{2-11} 
                                   & mAP           & ATR       & mAP         & ATR      & mAP            & ATR           & mAP            & ATR           & mAP           & ATR           \\ \hline \hline
         Clean                     & 0.67          & --        & 0.77        & --       & 0.76           & --            & 0.71           & --            & 0.77          & --            \\ \hline
         DAG \cite{xiao2018generating} & 0.65          & 0.19      & 0.75        & 0.16     & 0.74           & 0.16          & 0.60           & 1.00          & 0.76          & 0.08          \\ \hline
         R18-DCA                   & 0.10          & 1.00      & 0.62        & 0.23     & 0.65           & 0.17          & 0.61           & 0.17          & 0.72          & 0.08          \\ \hline
         DLA34-DCA                 & 0.50          & 0.28      & 0.07        & 1.00     & 0.62           & 0.2           & 0.53           & 0.28          & 0.67          & 0.14          \\ \hline
         R18-SCA                   & 0.31          & 1.00      & 0.62        & 0.36     & 0.61           & 0.37          & 0.55           & 0.42          & 0.70          & 0.17          \\ \hline
         DLA34-SCA                 & 0.42          & 0.90      & 0.41        & 1.00     & \textbf{0.53}  & \textbf{0.65} & \textbf{0.44}  & \textbf{0.82} & \textbf{0.62} & \textbf{0.42} \\ \hline
         \end{tabular}
 \end{center}
 
 \caption{Black-box attack results on the PascalVOC dataset. \textbf{From} and \textbf{To} means the same as Table \ref{tab3}.}
 \label{tab4}
\end{table*}

\begin{table*}[t]
 \centering
 \resizebox{1.35\columnwidth}{!}{
 \begin{tabular}{l|l||l|c|c|c|c}
 \hline
                    & Method& Network         & Clean       & Attack     & ASR (\%)       & Time (s)             \\ \hline \hline
 \multirow{6}{*}{\rotatebox[origin=c]{90}{PascalVOC}}&DAG~\cite{xie2017adversarial} & FR  & 0.70   & 0.050   & 0.92 & ~~9.8  \\ 
 &UEA~\cite{wei2018transferable}   & FR    & 0.70 & 0.050 & 0.93          & --     \\ 
 &SCA                              & R18   & 0.67 & 0.060 & 0.91          & ~20.1  \\ 
 &SCA                              & DLA34 & 0.77 & 0.110 & 0.86          & ~91.5  \\ 
 &DCA                              & R18   & 0.67 & 0.070 & 0.90          & ~~~0.3 \\
 &DCA                              & DLA34 & 0.77 & 0.050 & \textbf{0.94} & ~~~0.7 \\ \hline
 
 \multirow{4}{*}{\rotatebox[origin=c]{90}{MS-COCO}}
 &SCA       & R18   & 0.29  & 0.027 & 0.91          & ~50.4\\ 
 &SCA       & DLA34 & 0.37  & 0.030 & 0.92          & 216.0\\ 
 &DCA       & R18   & 0.29  & 0.002 & 0.99          & ~~~~\textbf{1.5}\\ 
 &DCA       & DLA34 & 0.37  & 0.002 & \textbf{0.99} & ~~~~2.4  \\ \hline
 \end{tabular}
 }
 \vspace*{0.1cm}
 \caption{Overall white-box performance comparison. The top row denotes the metrics. Clean and Attack denotes the mAP of clean input and adversarial examples. Time is the average time to generate an adversarial example.}
 \label{overall_white}
 \vspace*{-0.5cm}
\end{table*}

\noindent \textbf{PascalVOC/MS-COCO.} We evaluate our method on  two popular object 
detection benchmarks: PascalVOC \cite{everingham2015pascal} and MS-COCO~\cite{lin2014microsoft}. 

For the PascalVOC dataset, we follow previous research~\cite{xie2017adversarial,zhou2019objects} to split it to training/testing split: the trainval sets of PascalVOC 2007 and PascalVOC 
2012 are used as the training set to train all object detection models, and our adversarial examples are generated on the test set of PascalVOC 2007, which consists of 4,592 images. Both PascalVOC 2007 and PascalVOC 2012 has 20 categories. 

For the MS-COCO dataset, we train the object detection models with the training set of MS-COCO 2017 which includes 118,000 images, and the test set consists of 5,000 images. The MS-COCO 2017 dataset contains 80 object categories.

\noindent \textbf{Evaluation Metrics.} We use three metrics for the attacking performance of the generated adversarial examples:

i) \textbf{Attack Success Rate (ASR)}: ASR measures the attack success rate of the generated adversarial examples. It is computed by calculating the mAP (mean Average Precision) loss when substituting the original input images with the generated adversarial examples. 
\begin{equation}
   \begin{aligned}
       ASR = 1-\frac{mAP_{attack}}{mAP_{clean}}
\end{aligned}
\end{equation}
where $mAP_{attack}$ denotes the $mAP$ of detector with adversarial example, $mAP_{attack}$ 
denotes the mAP of clean input.

ii) \textbf{Attack Transfer Ratio (ATR)}: Transferability is an important property of the adversarial examples~\cite{papernot2017practical}. We evaluate the transferability by computing $ATR$:
\begin{equation}
  \begin{aligned}
      ATR = \frac{ASR_{target}}{ASR_{origin}}
\end{aligned}
\end{equation}
where $\mathop{ASR_{target}}$ represents the $\mathop{ASR}$ of the attacked target model and $\mathop{ASR_{origin}}$ denotes the $\mathop{ASR}$ of the model which generate the adversarial example. Higher ATR denotes better transferability.

iii) \textbf{Perceptibility}: 
The adversarial perturbation's perceptibility is quantified by its $L_p$ norms. Specifically, $P_{L_2}$ and $P_{L_0}$ are used, which are defined as follows.

\textbf{i) $P_{L_2}$:} $L_2$ norm of the perturbation. A lower $L_2$ value usually signifies that the perturbation is more imperceptible for the human. Formally, 
\begin{equation}
    P_{L_2} = \sqrt{\frac{1}{k}\sum{r_k^2}}
\end{equation}
where the $\mathop{k}$ is the number of the pixels. We also normalized the $\mathop{P_{L_2}}$ in $\mathop{[0,1]}$. 

\textbf{ii) $P_{L_0}$:} $L_0$ norm of the perturbation. A lower $L_0$ value means that fewer pixels are changed during the attack. We compute $P_{L_0}$ by measuring the proportion of changed pixels.

\noindent \textbf{White-Box/Black-Box Attacks.} We conducted both white-box and black-box attacks on several popular object detection models:

i) \textbf{White-Box Attack:} We conducted attack experiments on two models. Both models use CenterNet but with different backbones: Resdcn18 \cite{he2016deep} and DLA34 \cite{yu2018deep}. We trained these two models on PascalVOC and MS-COCO respectively. We obtain four models from these two backbones and two datasets. We evaluate the attack performance of SCA and DCA on each model. In this paper, we denote R18 as CenterNet with Resdcn18 backbone and DLA34 as CenterNet with DLA34 backbone.

ii) \textbf{Black-Box Attack:}  We classify the black-box attack methods into two categories: cross-backbone and cross-network. In the cross-backbone attack, we evaluate the transferability with Resdcn101\cite{he2016deep} on PascalVOC and MS-COCO. In the cross-network attack, we use not only anchor-free object detector, but also two-stage anchor-based detectors: CornerNet~\cite{law2019cornernet}, Faster-RCNN~\cite{ren2015faster} and SSD300~\cite{liu2016ssd}. Faster-RCNN and SSD300 are only tested on PascalVOC. CornerNet is only tested on the MS-COCO. The backbone is Hourglass~\cite{newell2016stacked}. In this paper, we denote FR as Faster-RCNN with VGG16~\cite{simonyan2014very} backbone.

\noindent \textbf{Quantitative Analysis} We provide the perceptibility result of our methods in \ref{sec: perceptibility}. We also provide the result of different pixel clipped value $\epsilon_S$ of SCA in \ref{sec: pixel_clip_value}.

\noindent \textbf{Implementation Details.} For both white-box and black-box attacks, we clip the range of the pixel values of the adversarial examples to $[0,255]$.
The attack threshold $t_{attack}$ in all experiments set to $0.1$.

\subsection{White-Box Attack Result}
\noindent

We first compare the white-box attack results of our method with the state-of-the-art 
attack algorithms on PascalVOC.  The attack results on CenterNet are summarized in 
Table~\ref{overall_white}. From the top half of Table~\ref{overall_white}, we find that: 
\textbf{(1)} DCA achieve higher ASR than DAG and UEA and the SCA achieve state-of-art attack performance.
\textbf{(2)} DCA is 14 times faster than the DAG. Because UEA is based on GAN, the total 
attack time should include the training time, but  UEA is not open source and we cannot 
confirm the training time, so we set the attack time of UEA as N$/$A.

\begin{figure*}[t]
 \centering
 \includegraphics[width=1\linewidth, height=0.9\textheight]{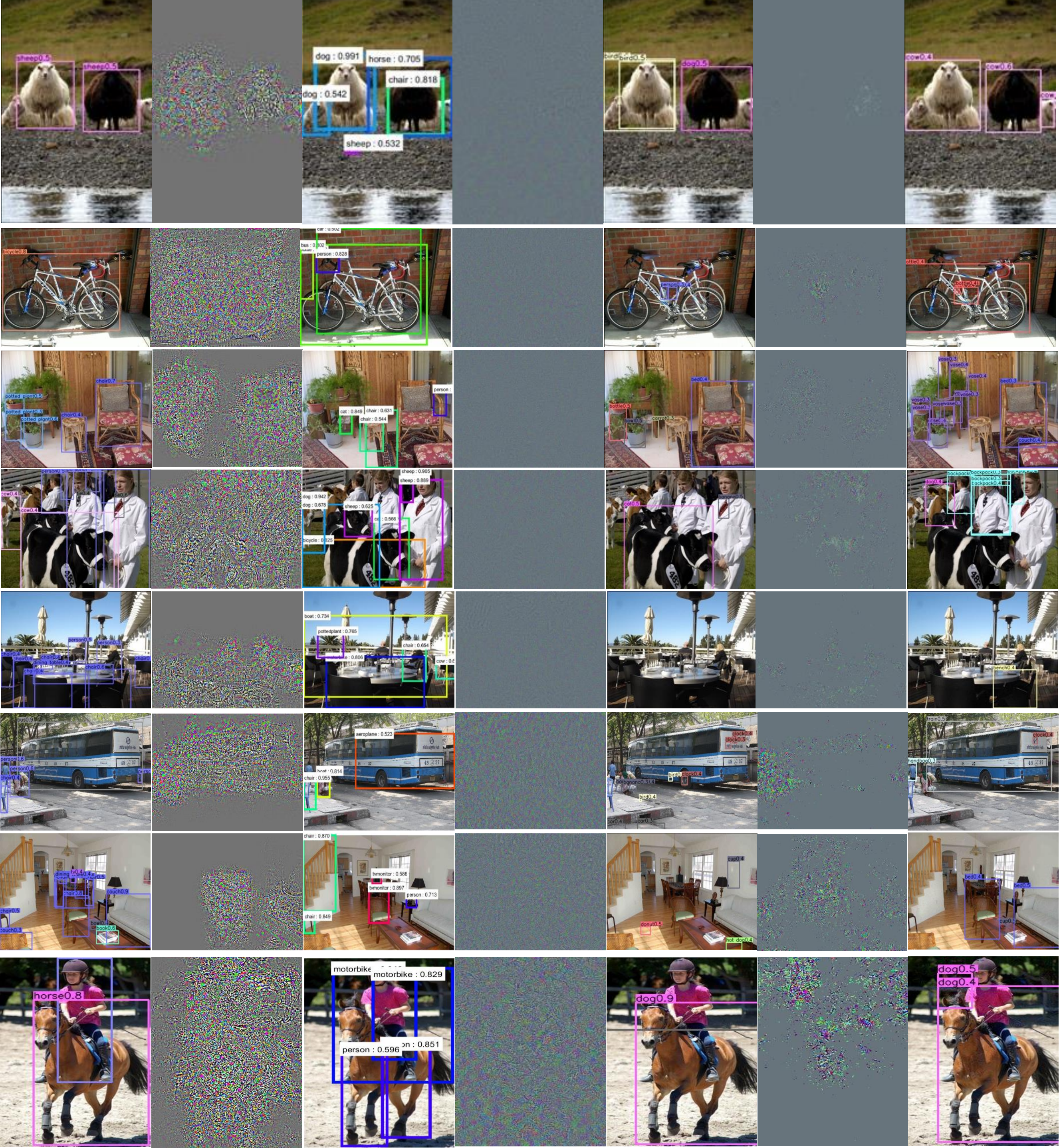}
 \caption{Each row is an example. \textbf{Column 1:} Detection results of clean inputs on CenterNet. \textbf{Column 2$\&$3:} DAG perturbations and DAG attacked results on Faster-RCNN. \textbf{Column 4$\&$5:} DCA perturbations and DCA attacked results on CenterNet. \textbf{Column 6$\&$7:} SCA perturbations and SCA attacked results on CenterNet. Note that in \textbf{Column 6}, from top to bottom, the percentage of the changed pixels for each SCA perturbations are: 3.12\% 3.4\%, 3.51\%, 5.91\%, 3.35\%, 6.04\%, 11.72\%, 8.19\%. We can see that the perturbations of DCA and SCA are smaller than the DAG's. Notably, the proposed SCA only changes a few percentage of pixels. To better show the perturbation, we have multiplied the intensity of all perturbation images by 10.} 
 \label{fig: effi_framework}
\end{figure*}

\begin{figure*}[t]
    \centering
    \includegraphics[width=1\linewidth, height=0.88\textheight]{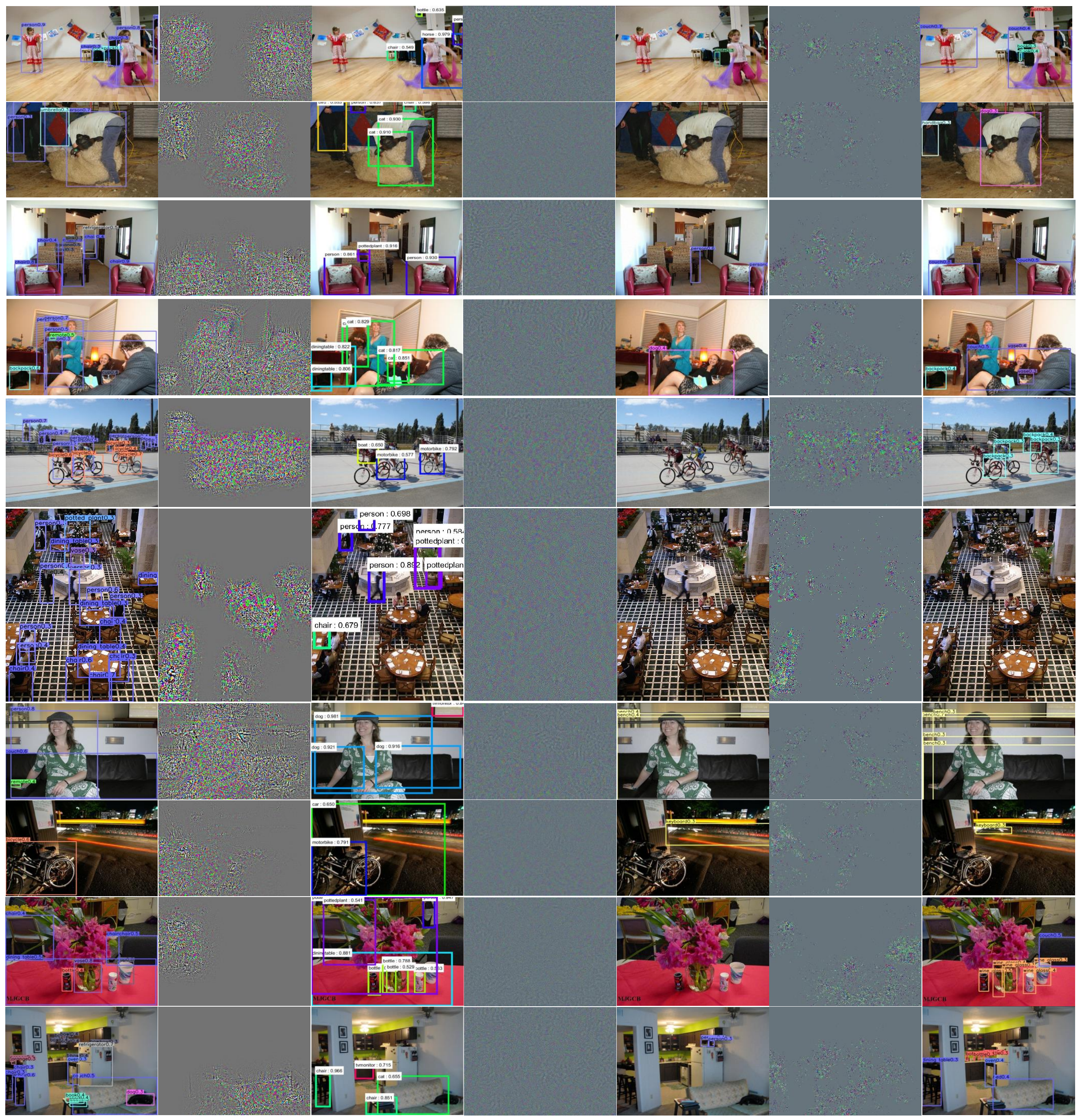}
    \caption{Each column is an example. \textbf{Column 1:} Detection results of clean inputs on CenterNet. \textbf{Column 2$\&$3:} DAG perturbations and DAG attacked results on Faster-RCNN. \textbf{Column 4$\&$5:} DCA perturbations and DCA attacked results on CenterNet. \textbf{Column 6$\&$7:} SCA perturbations and SCA attacked results on CenterNet. Note that in \textbf{Column 6}, from top to bottom, the percentage of the changed pixels for each SCA perturbations are: 3.75\%, 2.26\%, 5.99\%, 3.97\%, 9.12\%, 3.97\%, 2.00\%, 2.09\%, 3.93\%, 2.88\%. We can see that the perturbations of DCA and SCA are smaller than the DAG's. Notably, the proposed SCA only changes a few percentage of pixels. To better show the perturbation, we have multiplied the intensity of all perturbation images by 10.test} 
    \label{qc2}
\end{figure*}

The bottom half of Table~\ref{overall_white} shows the attack performance of our methods 
on MS-COCO. We only compare ASR with DAG and UEA as they didn't report their performance 
on MS-COCO. ASR of SCA on two backbones achieve almost the same ASR as the DAG and UEA on 
PascalVOC: 92.8\% and the ASR of DCA is higher than 99\%, meaning that almost all objects can not be detected correctly. It is obvious that both DCA and SCA achieve state-of-the-art attack performance on different models. We also show some qualitative comparison results between DAG and our methods in Fig.~\ref{fig: effi_framework} and Fig.~\ref{qc2}.

\begin{figure*}[t]
   \begin{center}
   \includegraphics[width=0.49\linewidth]{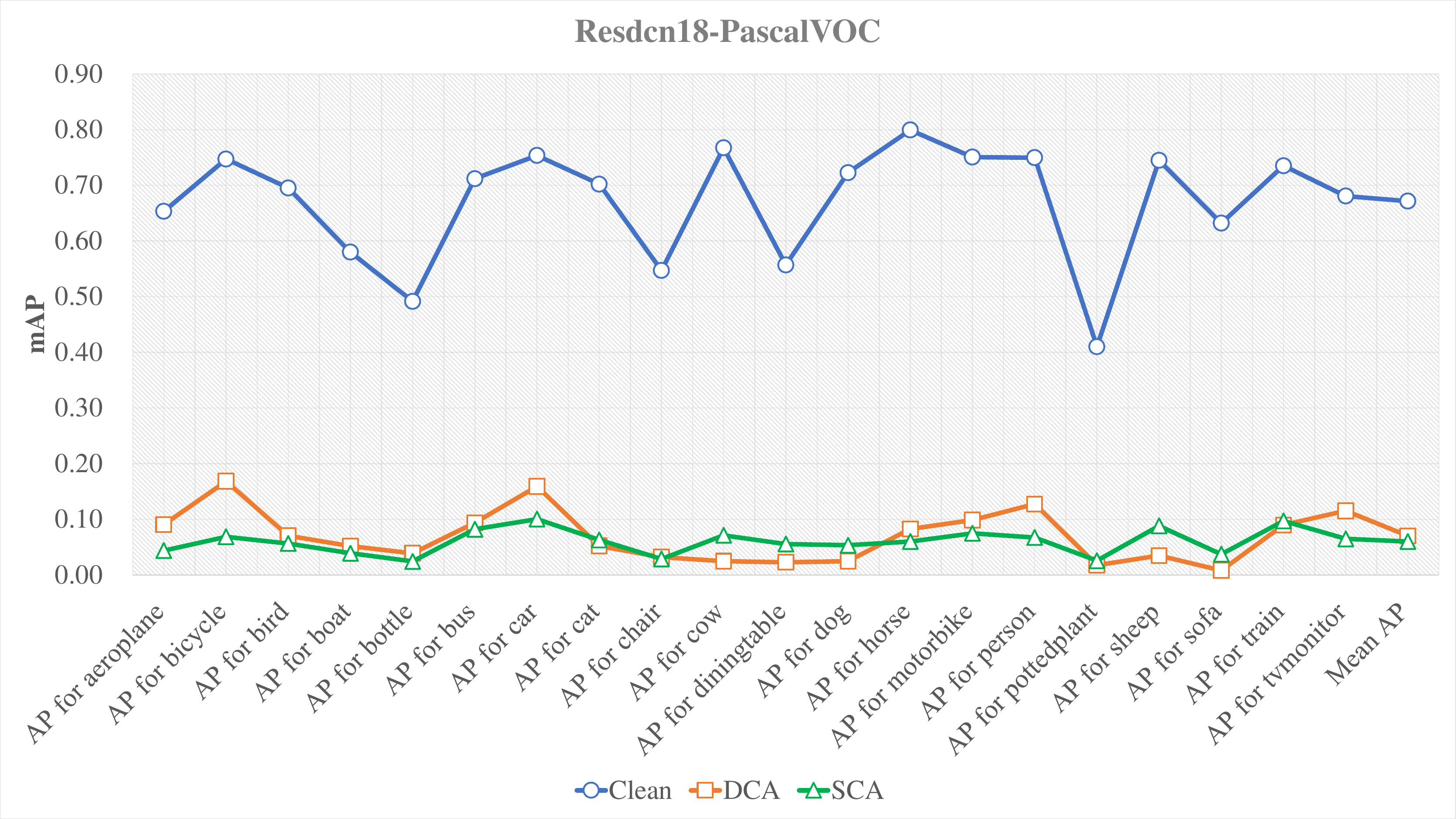}
   \includegraphics[width=0.49\linewidth]{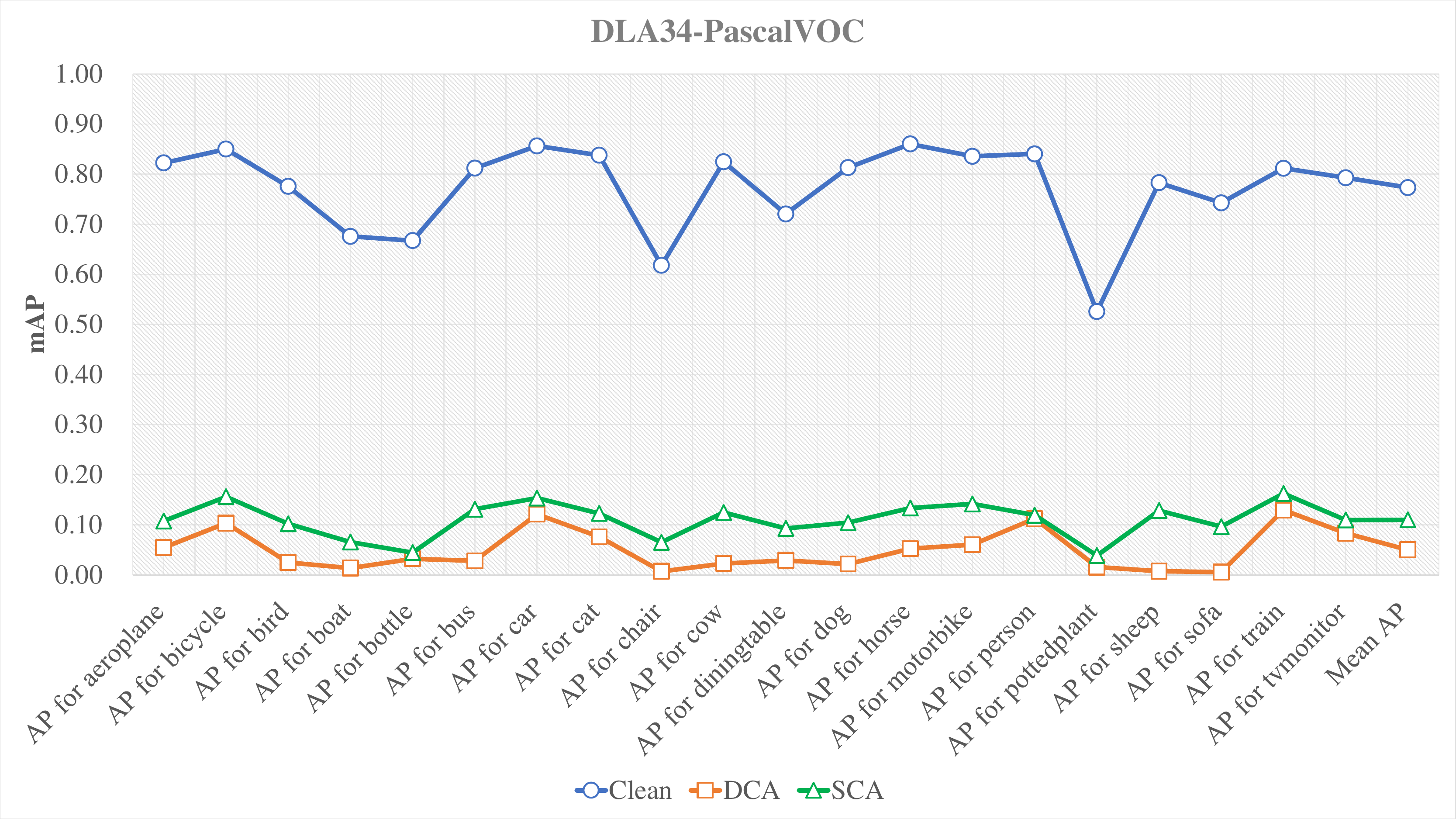}
   \caption{Quantitative analysis on PascalVOC.
   \textbf{Top:} The AP of each object category on the clean input and adversarial examples generated by CenterNet with Resdcn18 backbones on PascalVOC. \textbf{Bottom:} The AP of each object category of clean input and adversarial examples on DLA34.
   }
   \label{pascal_detail}
   \end{center}
\end{figure*}

\begin{figure*}[t]
   \begin{center}
   \includegraphics[width=0.49\linewidth]{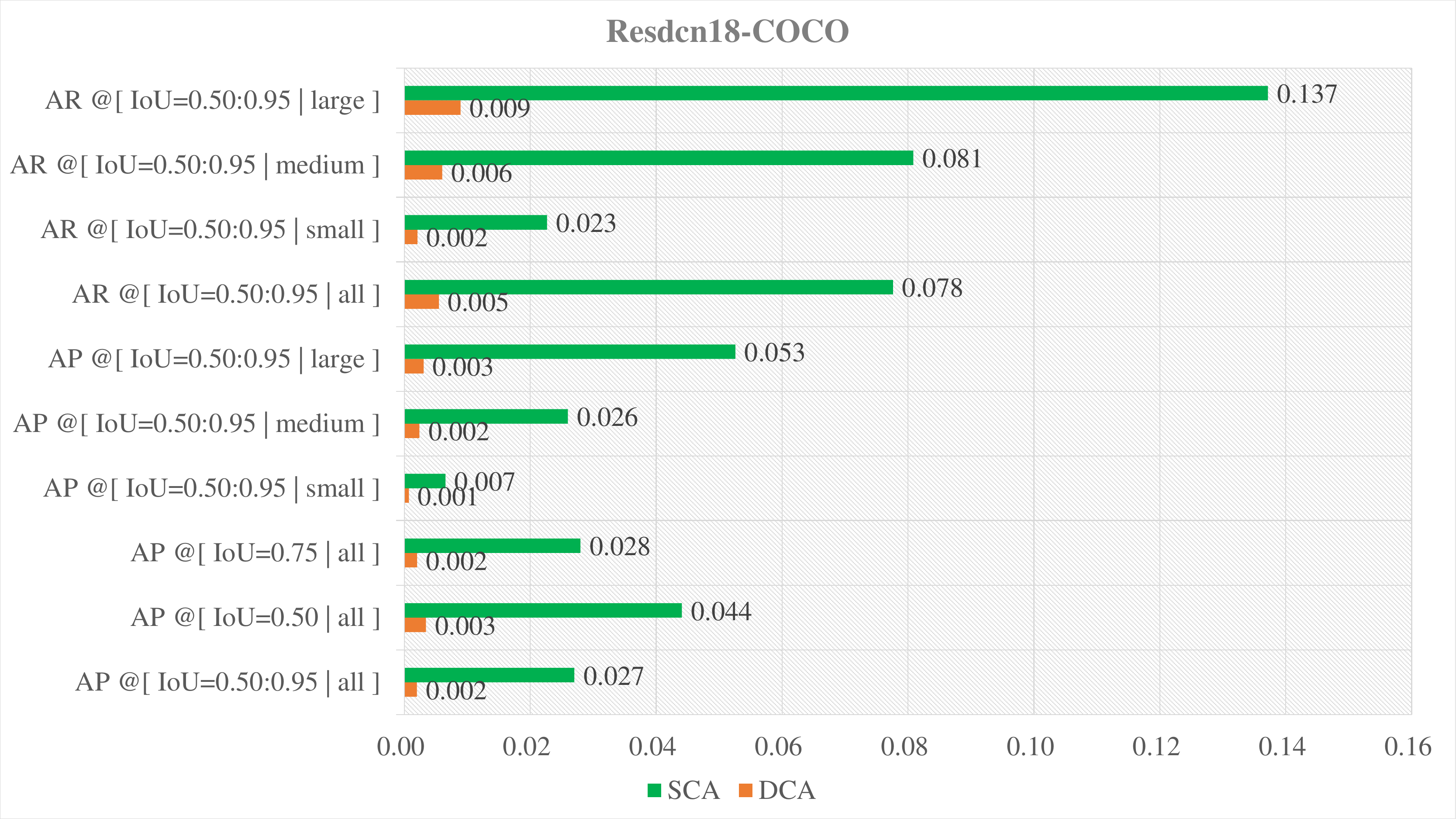}
   \includegraphics[width=0.49\linewidth]{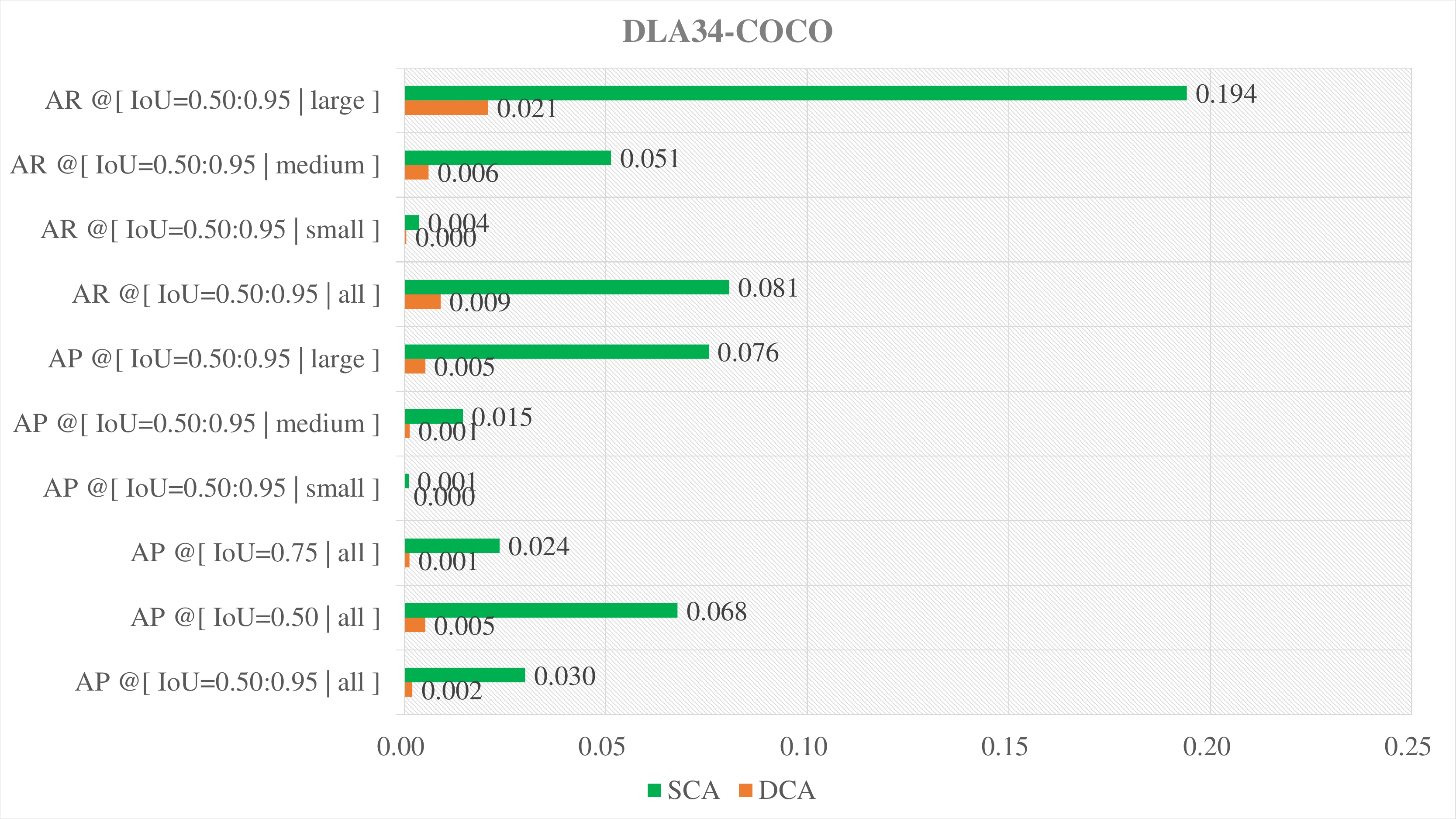}
   \caption{Quantitative analysis on MS-COCO.
   \textbf{Top:}  The AP and AR of SCA and DCA on MS-COCO. The backbone of  CenterNet is Resdcn18.
   \textbf{Bottom:} The AP and AR of SCA and DCAon MS-COCO. The backbone of CenterNet is  DLA3.
   }
   \label{coco_detail}
   \end{center}
\end{figure*}

Fig.~\ref{pascal_detail} shows the Average Precision (AP) of each object category on clean input and the adversarial examples generated by Centernet with Resdcn18 and DLA34 backbones. The AP drops roughly by the same percentage for all object categories. Fig.~\ref{coco_detail} shows the AP and Average Recall (AR) of SCA and DCA. We noticed that small objects are more vulnerable to adversarial examples than bigger ones. One possible explanation is that bigger objects usually have more key points than the smaller objects on the heatmap and our algorithm needs to attack all of them.

\subsection{Black-Box Attack Result/Transferability}
To simulate a real-world attack transferring scenario, we use DCA or SCA to generate the adversarial examples on the CenterNet and save them in the JPG format. Saving the adversarial examples in JPG format (JPG compression) may cause them to lose the ability to attack the target models~\cite{dziugaite2016study} as some key detailed information may be lost during the process. Then, we reload them to attack the target models and compute $mAP$. This puts a higher demand on the adversarial examples and further guarantees the transferability of the adversarial examples. 

\noindent \textbf{Attack transferability on PascalVOC.} The adversarial examples are generated on CenterNet with Resdcn18 and DLA34 backbones respectively. We use these adversarial examples to attack the other four models. All these five models are trained on the PascalVOC. We can find the DCA is more robust to JPG compression than SCA. But 
SCA achieves higher ATR than DCA in the black-box test. The results are summarized in Table~\ref{tab4}, which demonstrate that the generated adversarial examples can successfully transfer to CenterNet with other different backbones, including completely different types of object detectors: Faster-RCNN and SSD.

From Table~\ref{tab4}, we also find that DAG is sensitive to JPG compression, especially when the adversarial examples are used to attack Faster-RCNN. And adversarial examples generated by DAG basically lose the ability to attack CenterNet and SSD300. It means both DCA and SCA are better than DAG regarding transferability and the ability to resist JPG compression.

\noindent \textbf{Attack Transferability on MS-COCO.} As with PascalVOC, we generate adversarial examples on Centernet with Resdcn18 and DLA34 backbones and then use them to attack other object detection models. The results are summarized in Table~\ref{tab3}. 

The generated adversarial examples are not only can be used to attack other CenterNet with different backbones but also validity to attack CornerNet.

\clearpage

\clearpage

\begin{figure}[t]

    \centering
    \includegraphics[width=\linewidth]{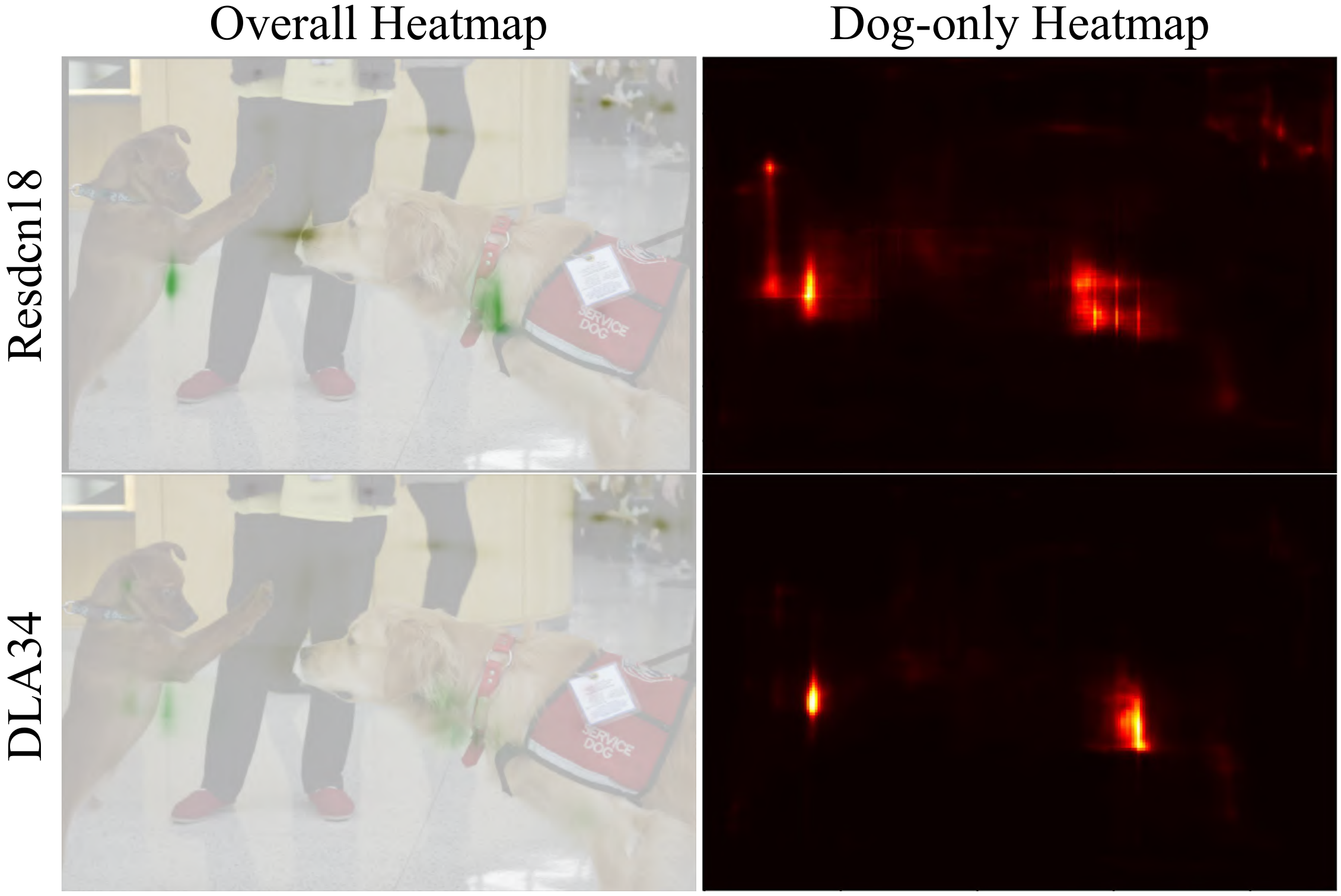}
    
    \centering
    \caption{Overall and Dog-only heatmaps generated by DLA34 and Resdcn18. The heatmap from DLA34 is more concentrated on the objects.}
    \label{hm_dog_compare}
\end{figure}

The overall performance of MS-COCO is higher than on PascalVOC. This is because the average $mAP$ of CenterNet on MS-COCO is lower than on PascalVOC. As a result, it makes the confidence of detection accuracies on MS-COCO lower than that on PascalVOC. Lower confidence means that the detector is weaker to defend against the adversarial examples. The adversarial example is robust to JPG compression on MS-COCO than on PascalVOC.

The adversarial examples generated on the DLA34 backbone achieve higher $\mathop{ASR}$ and $\mathop{ATP}$ than backbone Resdcn18, especially on MS-COCO. We also noticed that the adversarial examples generated by models with higher detection performance achieve better attack performance.
We guess this is because they produce more accurate heatmaps. To confirm this, we generate heatmaps for a randomly selected image with Centernet with DLA34 and Resdcn18 backbones (DLA34 achieves better detection performance). The heatmaps of the category dog are shown in Fig.~\ref{hm_dog_compare}. It demonstrates that the heatmap generated by DLA34 is more concentrated on the dogs in the image, while the heatmap predicted by Resdcn18 is more spread out.

\begin{table}[t]
   \centering
   \begin{tabular}{l||l|l}
   \hline
   Network & $\mathop{p_{l2}}$  & $\mathop{p_{l0}}$ \\ \hline \hline
   DAG & $\mathop{2.8 \times 10^{-3} }$ & $\mathop{\geq\ 99\%}$ \\ 
   R18-Pascal & $\mathop{5.1\times 10^{-3}}{(DCA)}$ & $\mathop{0.22\%} {(SCA)}$ \\
   DLA34-Pascal    & $\mathop{5.1\times 10^{-3}}{(DCA)}$ & $\mathop{0.27\%}{(SCA)}$ \\ 
   R18-COCO   & $\mathop{4.8\times 10^{-3}}{(DCA)}$ & $\mathop{0.39\%}{(SCA)}$ \\
   DLA34-COCO      & $\mathop{5.2\times 10^{-3}}{(DCA)}$ & $\mathop{0.65\%}{(SCA)}$ \\ \hline
\end{tabular}
\vspace*{0.1cm}
\caption{Quantitative perceptibility results of the generated perturbation.}
\label{qa}
\vspace{-0.3cm}
\end{table}

\subsection{Perceptibility}
\label{sec: perceptibility}
The perceptibility of the adversarial perturbations of DCA and SCA are shown on Table~\ref{qa}. We evaluate the perceptibility of DCA by $\mathop{p_{l2}}$ and $\mathop{p_{l0}}$. $\mathop{p_{l0}}$ of SCA is lower than 1\%, meaning that SCA can change only a few pixels to fool the detectors. It demonstrates that SCA generates sparse adversarial perturbations.

Although the average $\mathop{p_{l2}}$ of DCA is higher than DAG, the perturbations generated by DCA are still hard to be distinguished by human. The $\mathop{p_{l0}}$ of SCA is lower than $1\%$, meaning that SCA is successful to generate sparse adversarial perturbation. The $\mathop{p_{l0}}$ of DAG and DCA are higher than 99\%, far more than SCA.

\begin{figure}[t]
    \begin{center}
    \includegraphics[width=0.9\linewidth]{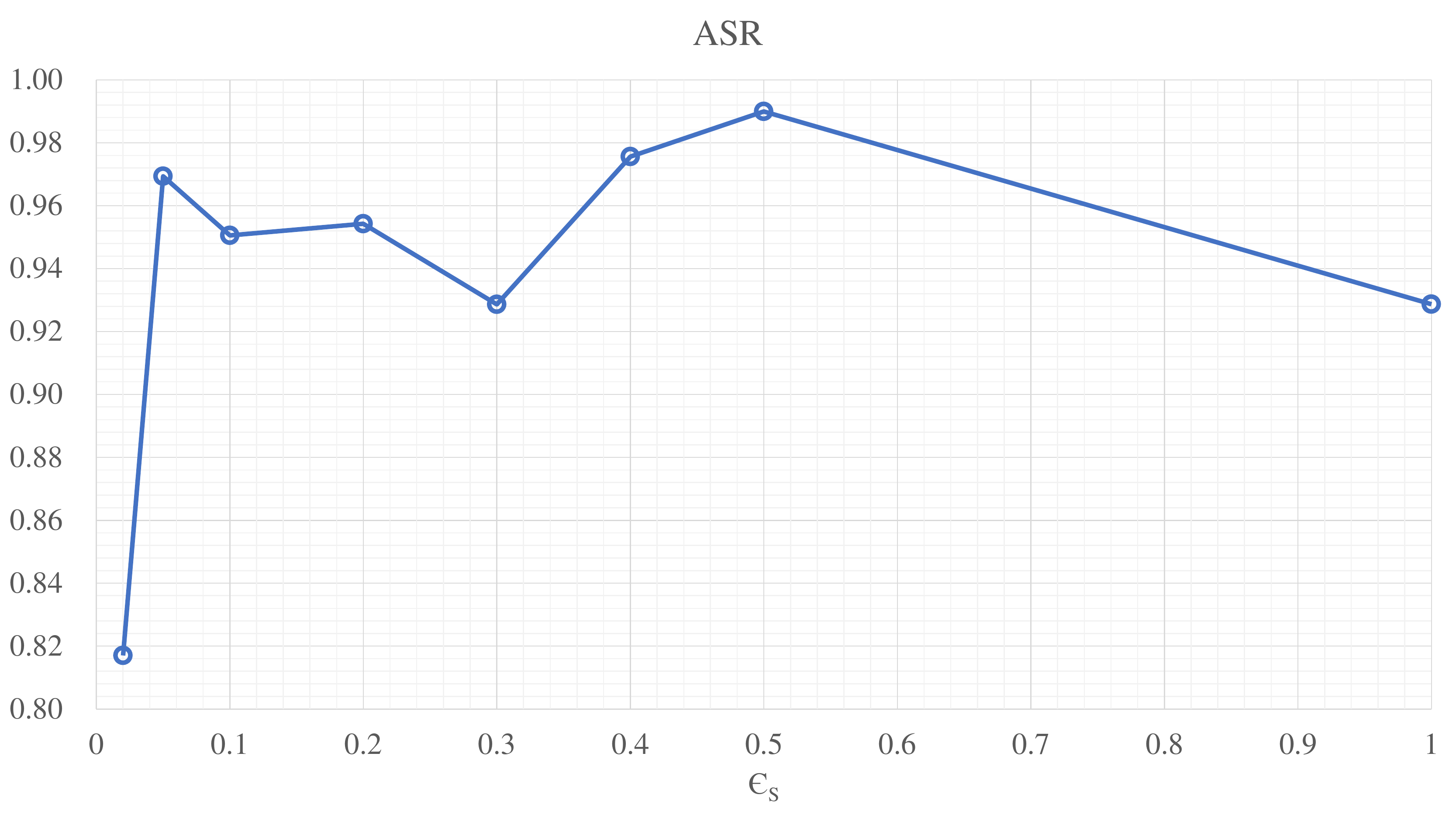}
    \includegraphics[width=0.9\linewidth]{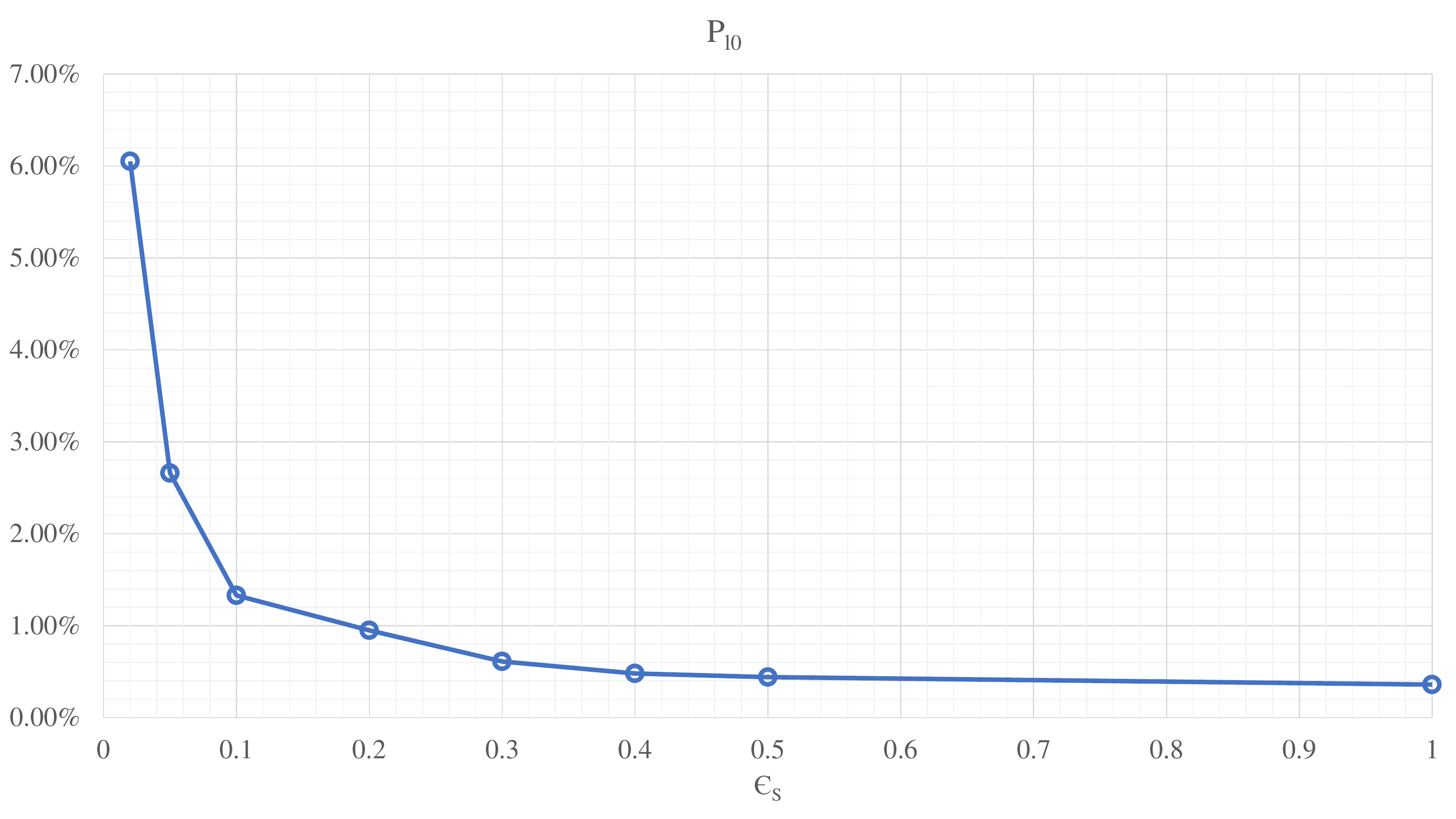}
    \includegraphics[width=0.9\linewidth]{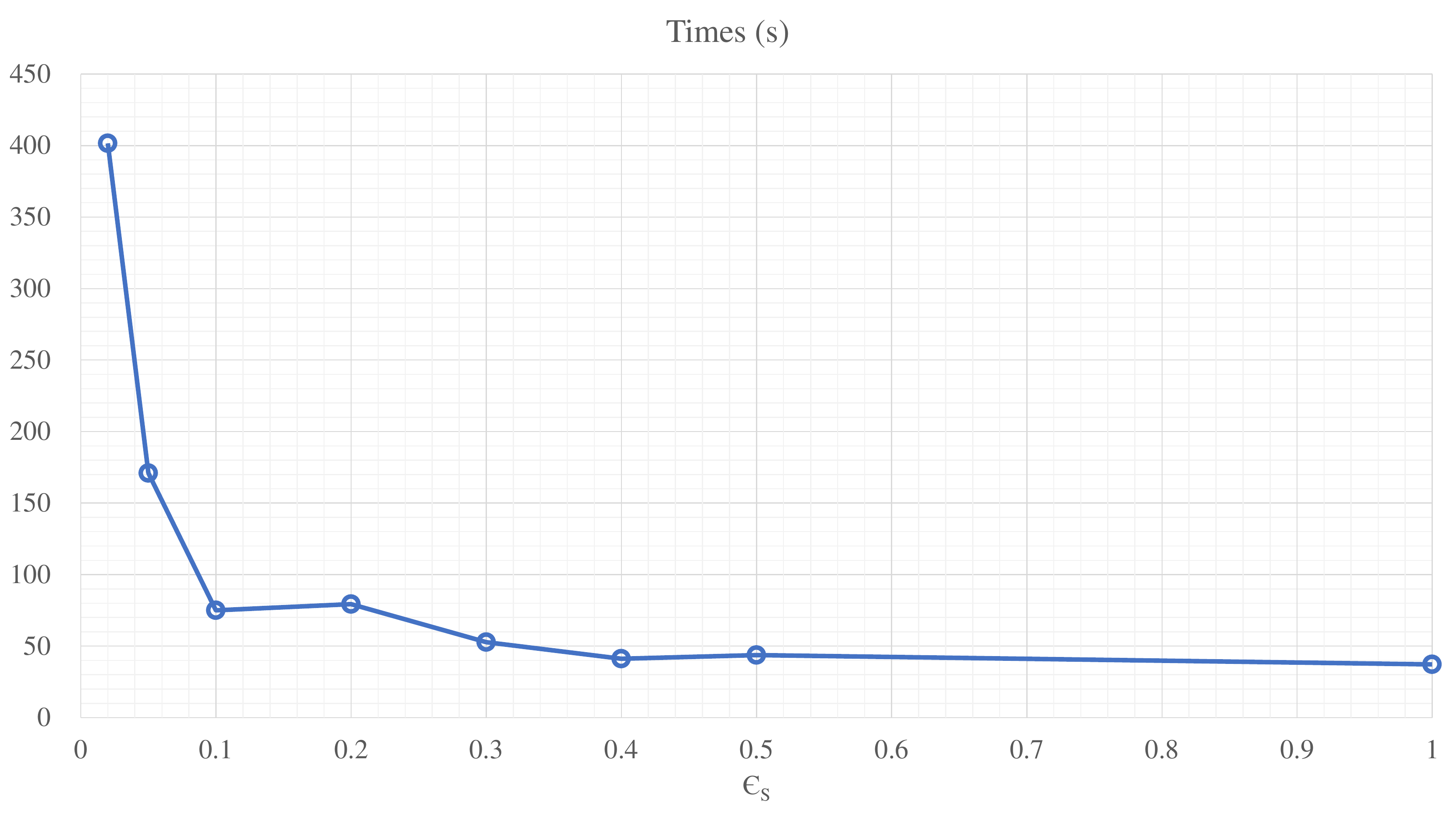}
    \caption{The connection between $\epsilon_S$ and Perturbation:
    \textbf{Top:} The ASR of SCA perturbation for different $\epsilon_S$.
    \textbf{Middle:} The $P_{l0}$ of SCA perturbation for different $\epsilon_S$.
    \textbf{Bottom:} The time required for SCA to generate perturbations for different $\epsilon_S$. The result shows higher $\epsilon_S$ will leading lower $P_{l0}$ and time consumption of our method. SCA achieves high white-box attack performance when $\epsilon_S$ higher than $0.05$. When $\epsilon_S$ lower than $0.05$, the white-box attack performance will be worse with a decrease of $\epsilon_S$. All results are tested on $100$ image samples of MS-COCO and  CenterNet with Resdcn18 backbone.
    }
    \label{epsilon}
    \end{center}
\end{figure}

\subsection{Pixel Clipped Value}
\noindent
\label{sec: pixel_clip_value}

In SCA, we clipping the intensity of perturbation to lie in the interval $x_i\pm\epsilon_S$ on each pixel of the input image $x_i$, so the perturbation is imperceptible to the human eyes. The results for different $\epsilon_S$ are shown in Fig.~\ref{epsilon}.

We can find that higher $\epsilon_S$ will lead to lower $P_{l0}$ and time consumption of our method. SCA achieves high white-box attack performance when $\epsilon_S$ higher than $0.05$. When $\epsilon_S$ lower than $0.05$, the white-box attack performance will be worse with a decrease of $\epsilon_S$. 

\section{Conclusion}
\label{sec: conclusion}
\noindent

In this paper, we propose two category-wise attack algorithms, SCA and DCA, for attacking anchor-free object detectors. Both SCA and DCA focus on global and high-level semantic information to generate adversarial perturbations. SCA is effectively in generating sparse perturbations by generating special decision boundary and approximating dense adversarial examples. DCA generates perturbations by computing gradients according to global valid categories.

Both SCA and DCA achieve state-of-the-art attack performance in the white-box attack experiments with CenterNet, where DCA is dozens of times faster than DAG and SCA while changing less than 1\% pixels. We also test the transferability of DCA and SCA under JPG compression. Both SCA and DCA achieve better attack transferring performance than DAG. The adversarial examples generated by our methods are less sensitive to JPG compression than DAG.

\section*{Acknowledgment}

This work was supported by the Sichuan Science and Technology Program under Grant 2019YFG0399.

\bibliographystyle{plain}
\bibliography{myref}

\end{document}